\documentclass[preprint,12pt]{elsarticle}

\usepackage{graphicx}
\usepackage{subfig}

\usepackage{amsmath}

\usepackage[colorlinks,linkcolor=blue]{hyperref}

\usepackage[top=2cm, bottom=2cm, left=2cm, right=2cm]{geometry}  

\makeatletter  
\newif\if@restonecol  
\makeatother

\usepackage[ruled,vlined,lined]{algorithm2e}
\usepackage{algpseudocode}

\usepackage{booktabs}
\usepackage{multirow}

\usepackage{lineno}
\usepackage{indentfirst}

\begin{document}
\begin{frontmatter}


\title{Carton dataset synthesis method for domain shift based on foreground texture decoupling and replacement}
\author[mymainaddress]{Lijun Gou}
\ead{getglj@hust.edu.cn}
\author[mymainaddress]{Shengkai Wu}
\author[mymainaddress]{Jinrong Yang}
\author[mymainaddress]{Hangcheng Yu}
\author[mymainaddress]{Chenxi Lin}
\author[mymainaddress]{Xiaoping Li\corref{mycorrespondingauthor}}
\cortext[mycorrespondingauthor]{Corresponding author}
\ead{lixiaoping@hust.edu.cn}
\author[mymainaddress]{Chao Deng}
\address[mymainaddress]{State Key Laboratory of Digital Manufacturing Equipment and Technology, Huazhong University of Science and Technology, Wuhan, 430074, China.}

\begin{abstract}
  One major impediment in rapidly deploying object detection models for industrial applications is the lack of large annotated datasets. We currently have presented the Sacked Carton Dataset(SCD) that contains carton images from three scenarios, such as comprehensive pharmaceutical logistics company(CPLC), e-commerce logistics company(ECLC), fruit market(FM). However, due to domain shift, the model trained with one of the three scenarios in SCD has poor generalization ability when applied to the rest scenarios. To solve this problem, a novel image synthesis method is proposed to replace the foreground texture of the source datasets with the texture of the target datasets. Our method can keep the context relationship of foreground objects and backgrounds unchanged and greatly augment the target datasets. We firstly propose a surface segmentation algorithm to achieve texture decoupling of each instance. Secondly, a contour reconstruction algorithm is proposed to keep the occlusion and truncation relationship of the instance unchanged. Finally, the Gaussian fusion algorithm is used to replace the foreground texture from the source datasets with the texture from the target datasets. The novel image synthesis method can largely boost AP by at least $4.3\%\sim6.5\%$ on RetinaNet and $3.4\%\sim6.8\%$ on Faster R-CNN for the target domain. Code is available \href{https://github.com/hustgetlijun/RCAN}{here}.
\end{abstract}
\end{frontmatter}
\section{Introduction}
\label{S:1}
  In the past few years, Convolutional Neural Networks (CNN) have significantly prompted the development of many computer vision tasks, such as image classification\cite{cen2021deep,mahmood2020resfeats}, object detection \cite{tong2020recent,zhang2019hyperfusion}, etc. Ideally, the training datasets for CNN-based models should satisfy the condition of independent identical distribution. But due to the change of lighting, imaging angle, and instance texture in different scenes, it's impossible for the datasets to strictly satisfy this condition, which causes the problem of domain shift. Thus, the CNN-based models trained with the datasets from the source domain generalize extremely badly on the datasets from the target domain.
  
\begin{figure}[htbp]
  \centering
  \subfloat[]{\label{Fig:2a}
  \includegraphics[width=0.3\linewidth]{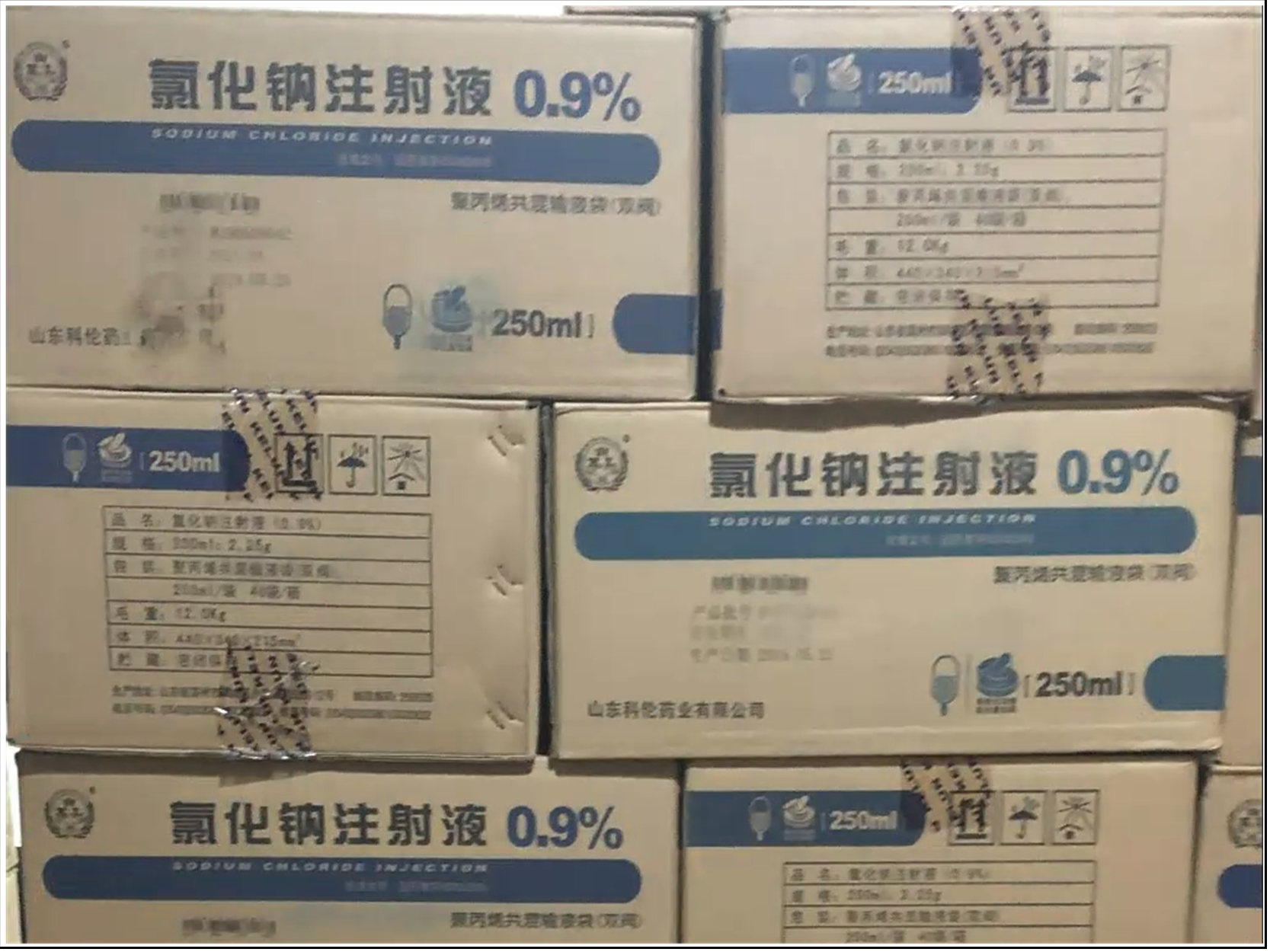}}
  \subfloat[]{\label{Fig:2b}
  \includegraphics[width=0.3\linewidth]{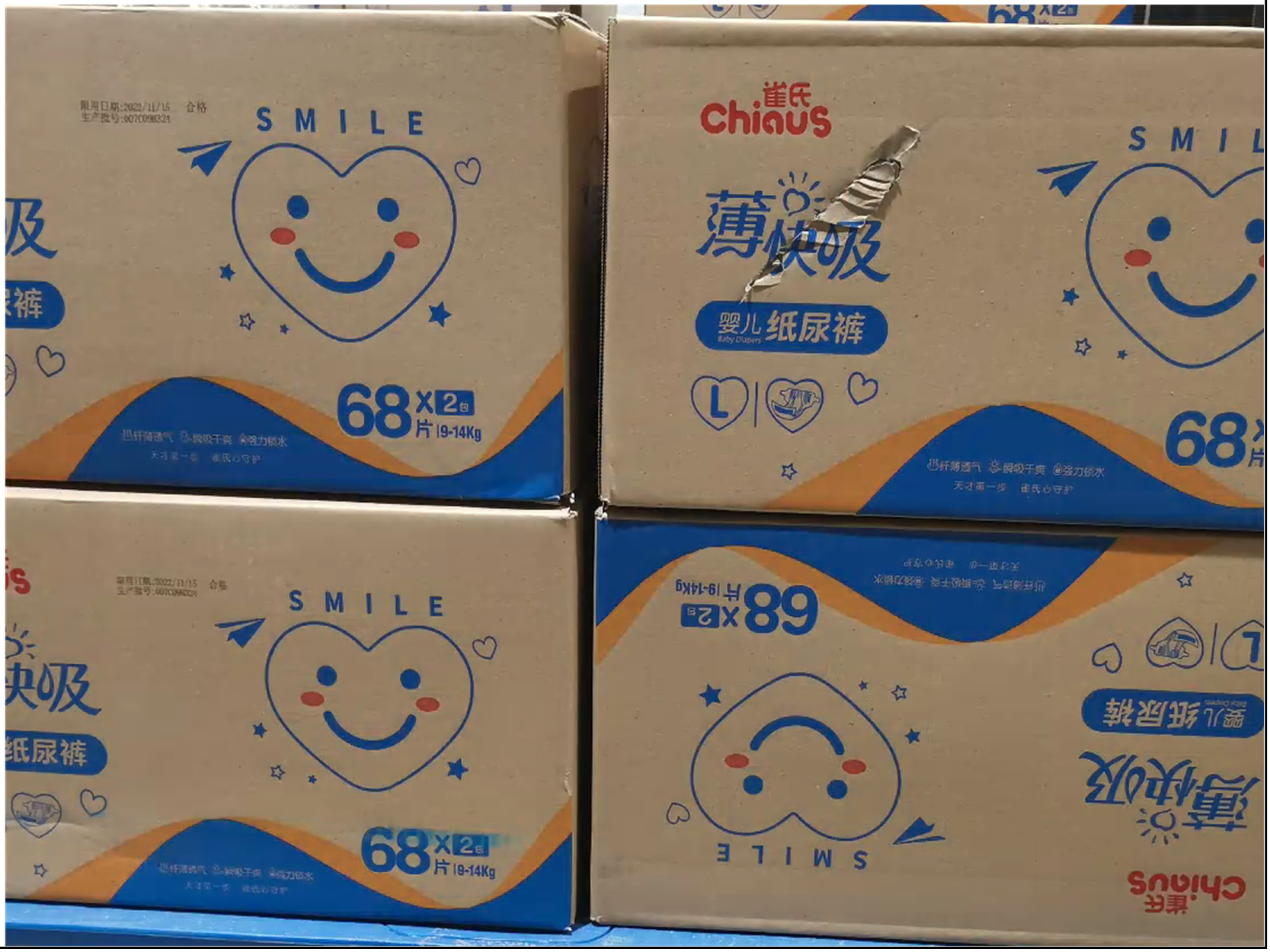}}
  
  \subfloat[]{\label{Fig:2c}
  \includegraphics[width=0.3\linewidth]{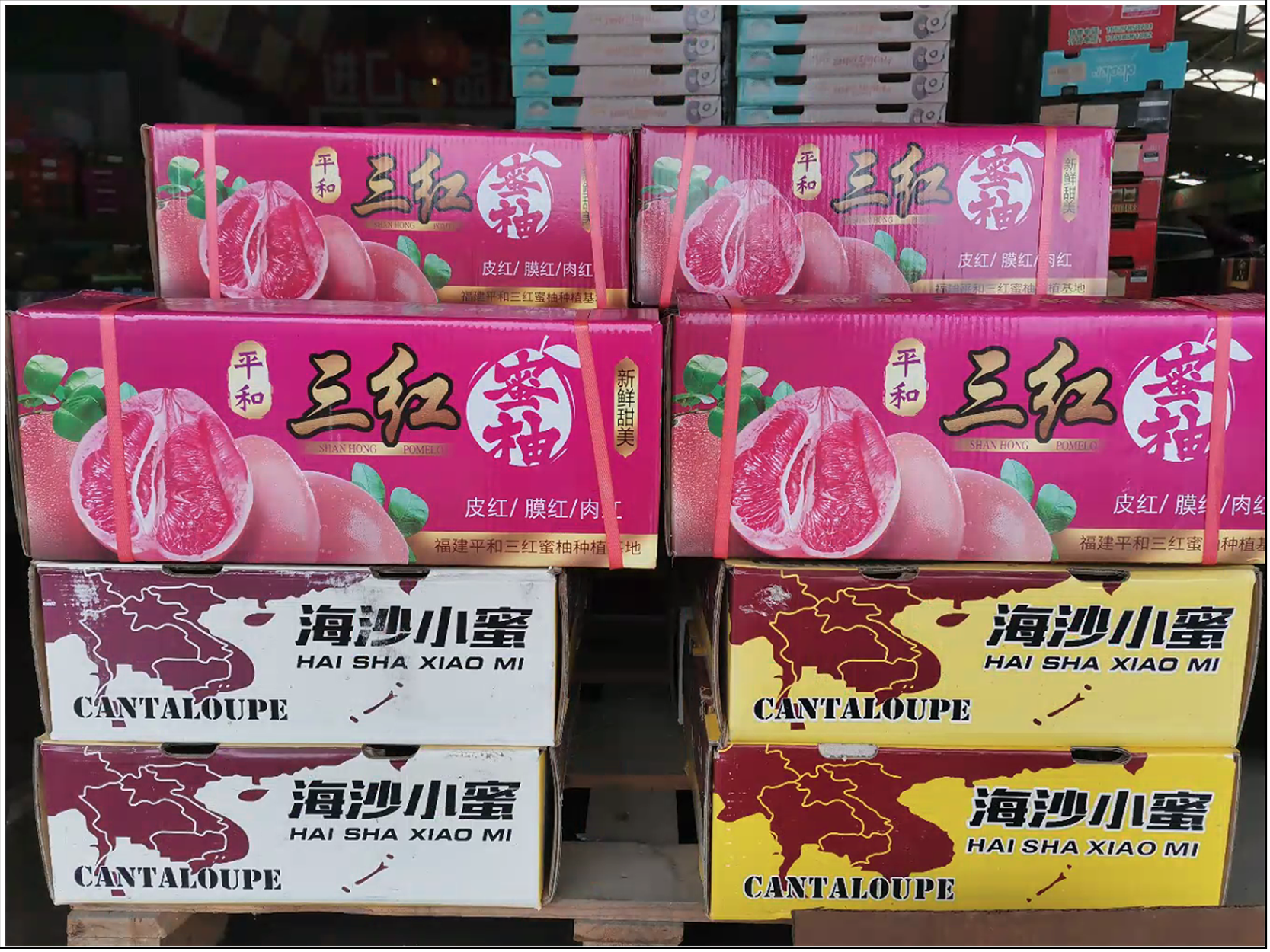}}
  \subfloat[]{\label{Fig:2d}
  \includegraphics[width=0.3\linewidth]{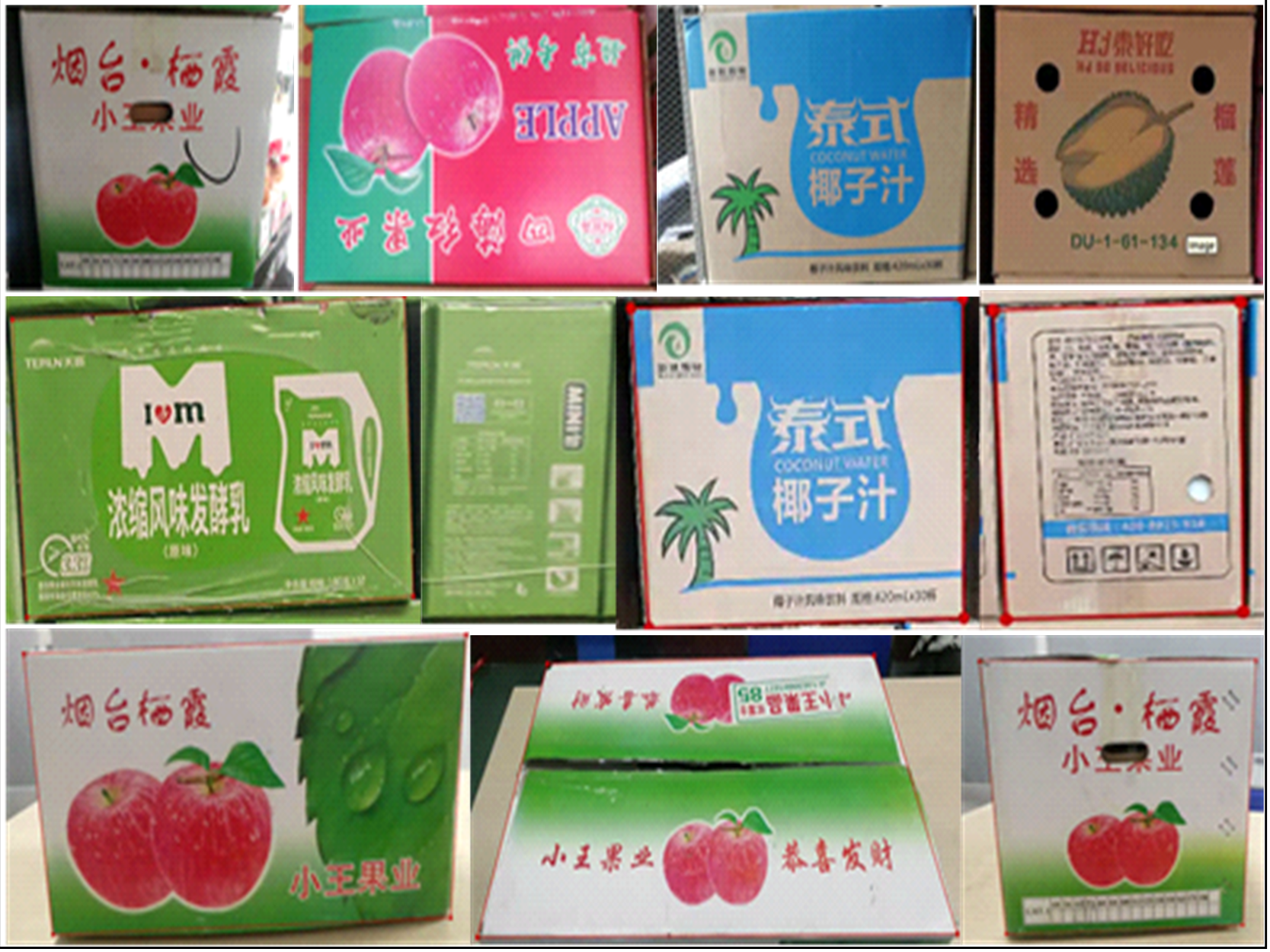}}
  \caption{(a) is the comprehensive pharmaceutical logistics company(CPLC); (b) is the e-commerce logistics company(ECLC); (c) is the fruit market(FM); (d) is the foreground texture dataset from FM.}
  \label{fig:2}
  \end{figure}

  The problem of domain shift is especially serious in the carton detection. With the development of the e-commerce logistics industry, carton detection becomes more and more important and can be applied to many logistics tasks such as carton sorting, destacking. SCD\cite{yang2021SCD} is a large scale carton dataset for carton detection and the carton images are collected from three different scenes. The carton images from different scenes distribute differently because of the inconsistency in the carton logo, color, texture, etc(as shown in Figure.\ref{fig:2}), which causes the serious problem of domain shift between the carton datasets from different scenes. Thus, the carton detection models trained with carton images from one scene(source domain) generalize badly to the carton images from other scenes(target domain).
  
\begin{figure}[htbp]
\centering
\includegraphics[height=0.3\linewidth]{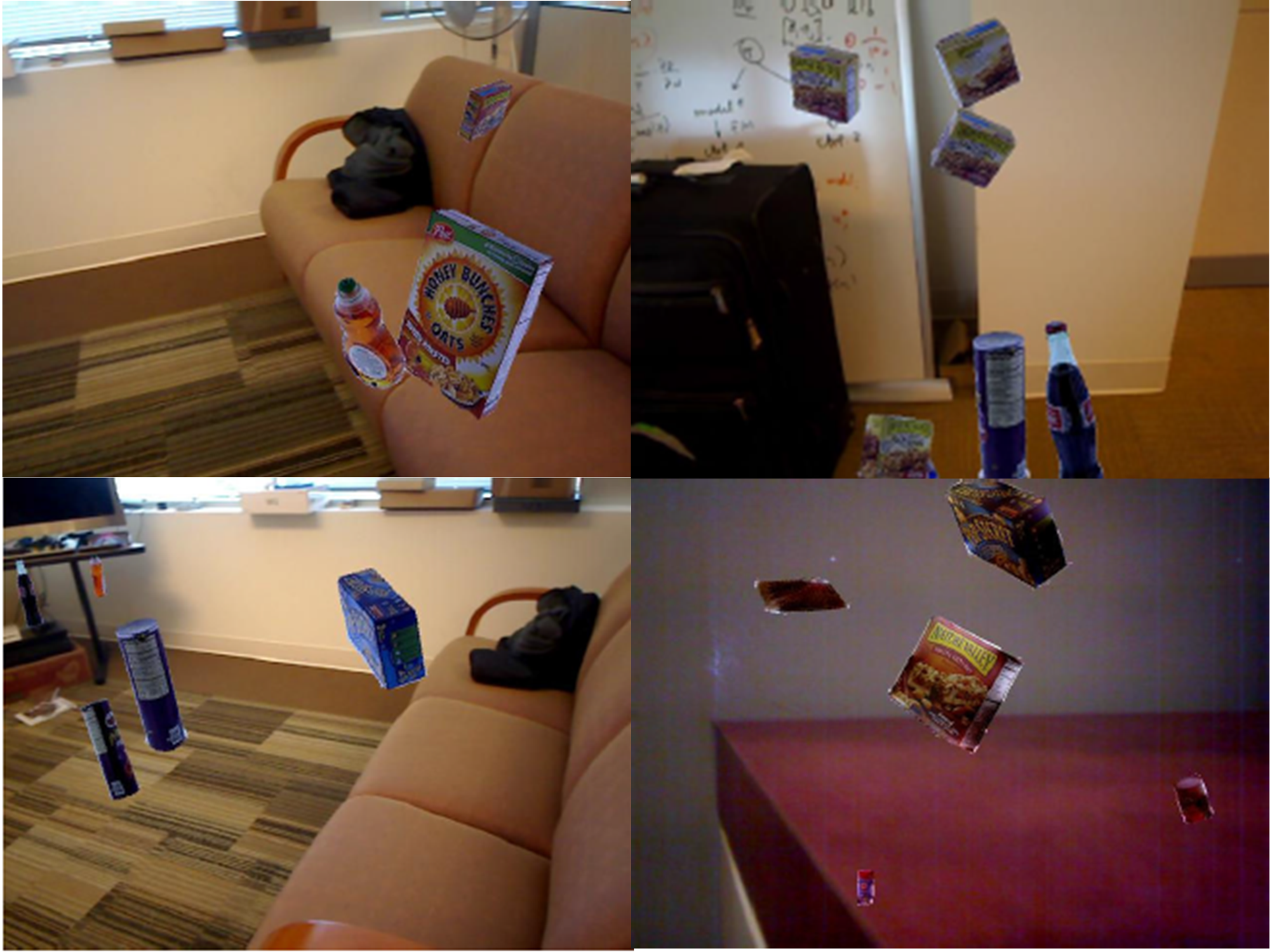}
\caption{Synthesized images of the cut-and-paste methods\cite{dwibedi2017cut}(600 images for each kind of object)}
\label{fig:1}
\end{figure}

  To solve the problem of domain shift, there currently exist three kinds of methods. (1) A large amount of data is collected and labeled to construct large-scale datasets, which is extremely expensive, time-consuming, and error-prone. Thus, the construction of large-scale datasets has become one of the bottlenecks for developing CNN-based models. (2) Data augmentation strategies(such as random cropping\cite{krizhevsky2017imagenet,szegedy2015going}, color jittering\cite{szegedy2015going}, etc) are proposed to increase the number of data by utilizing limited training data. However, these methods can only introduce limited data variation and can't solve the problem of domain shift well. (3) Crop-and-paste methods\cite{dwibedi2017cut,liu2020novel} are proposed to synthesize the training images, which crop the foregrounds with different poses and then randomly paste them on the backgrounds as shown in Fig.\ref{fig:1}. The context relationship between foregrounds and backgrounds is complex and important for recognizing the foregrounds\cite{tripathi2019learning}. However, these methods randomly crop and paste foregrounds without considering the context relationship between foregrounds and backgrounds. Thus, most of the synthesized images look unreal and hurt the generalization ability of the trained CNN models. 
  
   For solving the problem of domain shift in carton detection, it's expensive and difficult to extensively collect carton datasets for all the logistics scenes to obtain a carton detection models which generalize well to different scenes. How can we construct the carton datasets for the new scenes with a more simple method? In this paper, we propose a novel dataset synthesis method of replacing the foreground instance texture in source domain with the instance texture from the target domain to construct the carton datasets for the target domain. As shown in Fig.\ref{fig:3}, our method involves four key stages:(1) label the images with the proposed labeling rules; (2) segment different surface of the carton to decouple the carton textures; (3) construct a complete contour for the carton surface to preserve the occlusion and truncation relationship of the instance in backgrounds; (4) synthesize carton images with Gaussian fusion method to construct the carton datasets for the target domain. Compared with crop-and-paste methods\cite{dwibedi2017cut,liu2020novel}, our method can preserve the context relationship of foregrounds and backgrounds and the generated carton images look more real. In addition, 600 images for each kind of instance need to be collected in the crop-and-paste methods while only six surface textures for each carton need to be collected in our method, which is more simple and low-cost. Besides, our method can not only solve the problem of domain shift, but also serve as an excellent data augmentation strategy. Extensive experimental results on Faster R-CNN\cite{ren2016faster} and RetinaNet\cite{lin2017focal} demonstrate the effectiveness of our method on solving the problem of domain shift in the carton detection. 

   Our method has three main contributions: (1) A novel dataset synthesis method of replacing the foreground instance texture is proposed to generate the target dataset; (2) A surface segmentation method for texture decoupling is proposed to preserve the context relationship of foreground instance and backgrounds. (3) A contour reconstruction algorithm is proposed to keep the occlusion and truncation relationship of the instance unchanged in the real backgrounds.

\section{Related Work}
\label{S:2}
   
\subsection{Data Augmentations}
\label{subS:2.1}
  Data augmentations have played an key role in achieving state-of-the-art results for many computer vision tasks such as image classification\cite{cen2021deep,mahmood2020resfeats} on ImageNet\cite{krizhevsky2017imagenet}, object detection\cite{ren2016faster,tong2020recent,wu2020iou} on MS COCO\cite{lin2014microsoft}. The simplest strategy for solving domain shift is to train CNN-based models on a larger scale of data.  But it is difficult to build a large-scale dataset. Many data augmentation strategies have been proposed to augment the existing datasets such as random crop\cite{krizhevsky2017imagenet,szegedy2015going}, color jittering\cite{szegedy2015going}, perspective transformation\cite{huang2019faster}, Auto/RandAugment\cite{cubuk2018autoaugment,cubuk2020randaugment}, and random expansion\cite{ghiasi2020simple,liu2016ssd}. These augmentation strategies can effectively improve the performance of the source domain but can't solve the problem of domain shift well. So a method that can improve the performance of the source domain and solve the domain shift is important.

\subsection{Copy-Paste Augmentation}
\label{subS:2.2}
  The cut-and-paste methods\cite{georgakis2017synthesizing,dwibedi2017cut} have been used to enrich the datasets and improve the generalization ability of the model. At present, the cut-and-paste methods mainly focus on the contextual semantic relationship of the foreground instance and background image\cite{georgakis2017synthesizing,tripathi2019learning}, but the context relationship is complex, it is difficult to get the real context relationship. And these methods in Ref.\cite{dwibedi2017cut,tripathi2019learning,ghiasi2020simple} focus on how to ignore these subtle pixel artifacts during the synthesis process. Compared with the methods in Ref.\cite{dwibedi2017cut,liu2020novel,georgakis2017synthesizing,tripathi2019learning}, the difference of our method is that the scale and position of instance on the backgrounds remains unchanged and the occlusion and truncation relationship of the instance remains unchanged on the real backgrounds . And the carton instance is composed of multiple surfaces, different combinations of the texture can generate a new instance, so our method don't need to collect a large amount of foreground instances with different perspectives and poses only six texture images for each kind of carton.

\subsection{Domain adaptation}
\label{subS:2.3} 
  Domain adaptation\cite{ben2010theory,sanodiya2019novel,toldo2020unsupervised,zhang2020adversarial} aims to solve the problem of domain shift, which has been widely studied in many visual tasks\cite{guan2021scale,schutera2021cuepervision}. There are many methods to reduce domain shift such as feature distribution matching between source domains and target domains\cite{li2020structure,zheng2020cross,chen2018domain} based on adversarial learning\cite{ganin2016domain,rahman2020correlation}, and pix-to-pix style transformation\cite{hsu2020progressive,toldo2020unsupervised} based on cyc-GAN\cite{zhu2017unpaired}. And for the feature distribution matching method, there are two keys: Global feature distribution matching to reduce the differences between the source domain and target domain in brightness, viewing angle, etc, and local instance features distribution matching to reduce the differences of the instance in texture, scale, etc. But all the methods must be collecting large data from the target domain, it is time-consuming. And our method can quickly get a lot of data for target domain and it can solve the domain shift. Especially compared with the pix-to-pix style transformation method\cite{hsu2020progressive}, our method is stably.
\begin{figure}[htbp]
\centering
\includegraphics[height=0.45\linewidth]{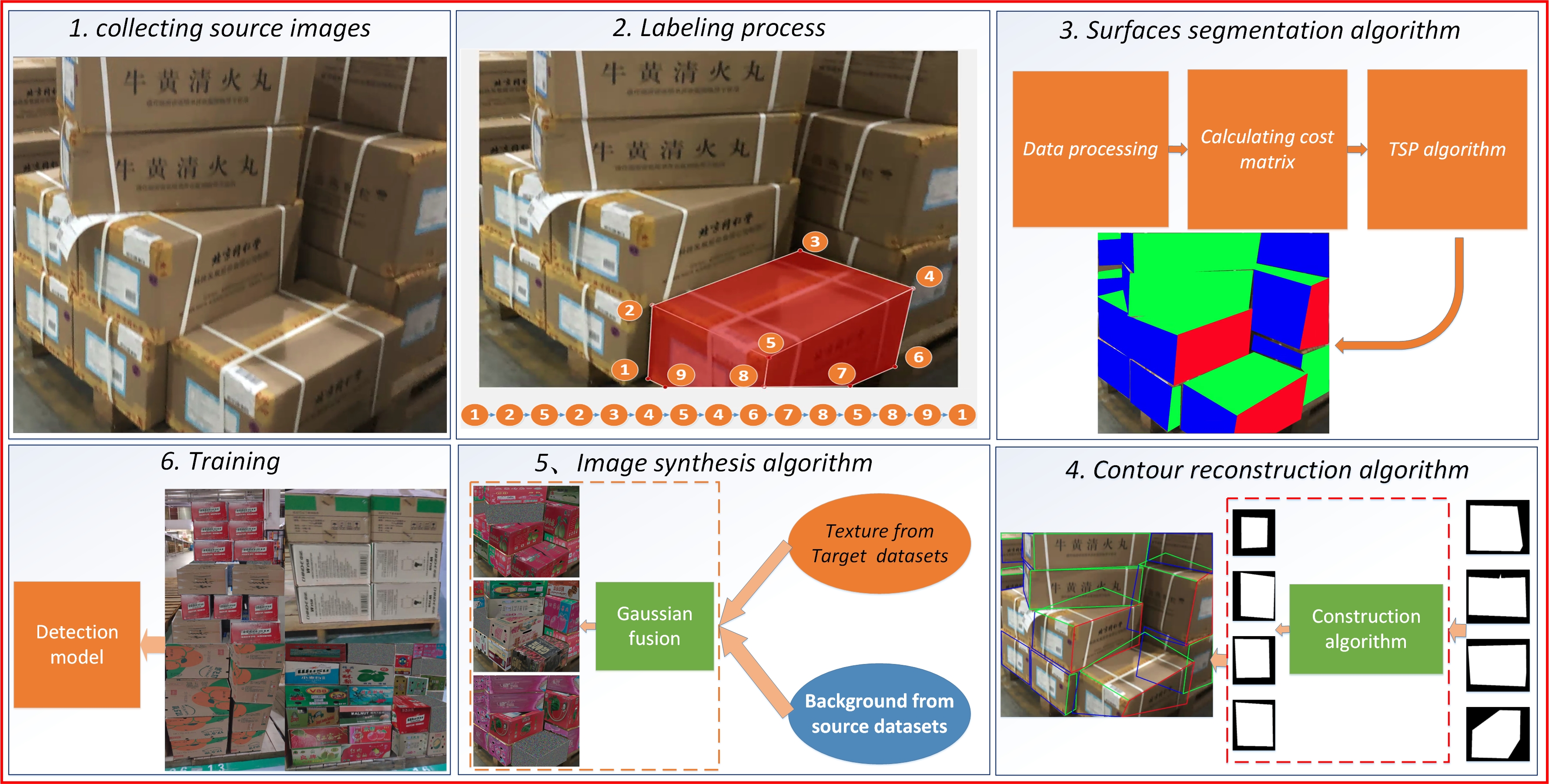}
\caption{The architecture of the novel carton datasets synthesis method.}
\label{fig:3}
\end{figure}
\section{Method}
\label{S:3}
\subsection{Approach overview}
\label{subS:3.1}
  We propose a simple approach to rapidly collect data for target domain with less time consumption and no human annotation. And the key idea of our approach is keeping the contour of the carton from source domain unchanging to replace their foreground texture with the target texture. This approach can capture all visual diversity of an instance with different views, scales, and orientation. And our approach can maintain the contextual semantic relationship of the image to make the synthesized image more realistic. The architecture of our carton datasets synthesis method is illustrated in Fig.\ref{fig:3}. And there are six steps of our method:
  
  (1) \textbf{Collect source images:} We have built a Stacked Carton Dataset(SCD)\cite{yang2021SCD} for carton detection task in the warehousing and logistics industry.
  
  (2) \textbf{Labeling process:} In the image plane, the carton may contain one, two, or three visible surfaces depending on the shooting angle and each surface has different texture. According to the traditional labeling rules, different surfaces cannot be distinguished. So we have designed new labeling rules(as shown in Sec.\ref{subS:3.2}) to help surface segmentation.
  
  (3) \textbf{Surfaces segmentation algorithm:} As shown in Fig.\ref{fig:label}, each surface of the carton is a texture and a closed polygon such as: 2-3-4-5, 4-6-7-8-5 and 1-2-5-8-9. So the surface segmentation is to find all the closed polygon without any overlap. Here, we consider the labeled point as a city, so the surface segmentation can be solved by the method of TSP\cite{gavish1978travelling}. Our algorithm consists of three steps(see Sec.\ref{subS:3.3} for detailed information) as shown in third module in Fig.\ref{fig:3}. The first step is data processing, the second step is calculating cost matrix as a directed graph and the third step is to get all closed polygons of the carton with the TSP algorithm to decouple the texture.
  
  (4) \textbf{Contour reconstruction algorithm:} In order to keep the occlusion and truncation relationship of the instance unchanged in the real backgrounds and the contour of the foreground texture from target datasets is complete as shown in Fig.\ref{Fig:2d}. So when the contour of the surface from source datasets is incomplete, we should construct a complete contour. And we construct a parallelogram as the complete contour for the surface(see Sec.\ref{subS:3.4} for detailed information). 
  
  (5) \textbf{Image synthesis:} Gaussian fusion method is used to generate images and the random nosie as negative texture example is used to make the detection model focus only on the object appearance(see Sec.\ref{subS:3.5} for detailed information).
  
  (6) \textbf{Training:} We use the generated images and the source images to train the detection models.

\subsection{Labeling method}
\label{subS:3.2}
  SCD\cite{yang2021SCD} mainly focus on the task of carton detection in the logistics industry. The images in SCD are collected from three scenarios of different locations. And each scenario contains a large number of neatly stacked cartons as shown in Figure \ref{fig:2}.

  \textbf{Data collection:} SCD mainly collects carton images in the loading and unloading dock scenes. Because the cameras of the palletizing robots is approximately parallel to the surface of the goods, the imaging plane is approximately parallel to the surface of the goods during data collection(as shown in Figure \ref{fig:2}). And the shooting distance is within 5 m.
\begin{figure}[]
\centering
\includegraphics[height=0.3\linewidth]{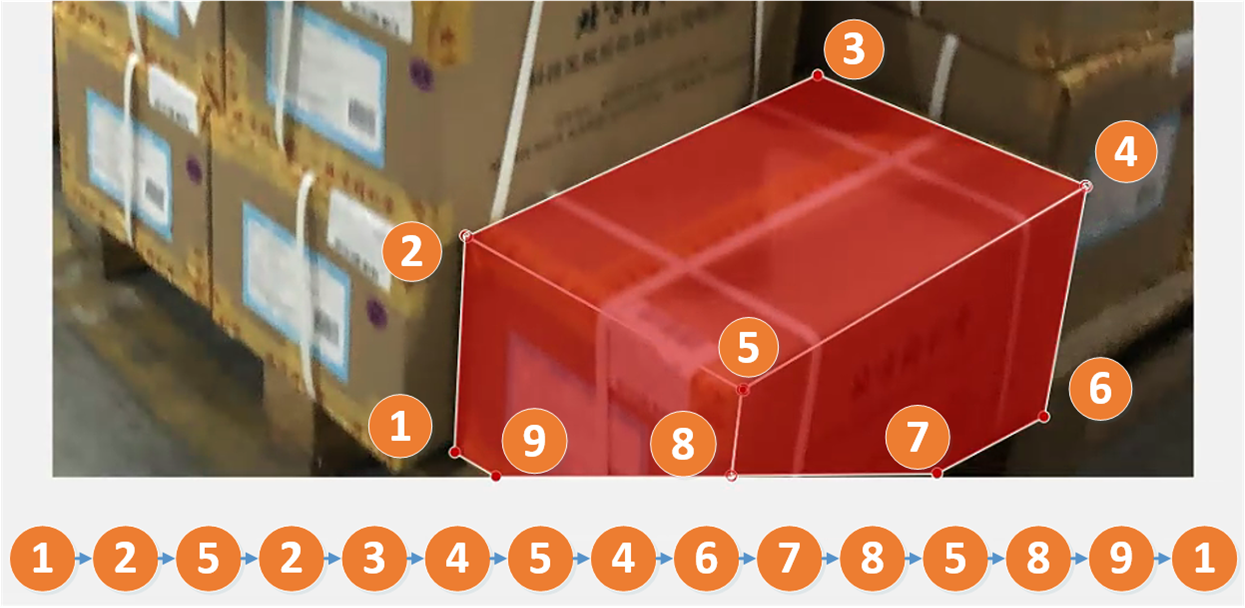}
\caption{Labeling method. The number represents labeled point and the arrows represents the labeling process. \textbf{Common lines:} the line between two visible surfaces, such as line 5-2, 5-4, 5-8. \textbf{Three-line points}: the points which are intersected by three visible lines such as point 5, 2, 4, 8; \textbf{Two-line points}: the points which are intersected by two visible lines such as point 1, 3, 6, 7, 9.}
\label{fig:label}
\end{figure}

  \textbf{Labeling rules:} SCD utilizes LabelMe\cite{russell2008labelme} for labeling. Besides, to help the robot pick up the goods, one auxiliary labling rule is proposed. \textbf{Occlusion/All:} it is labeled as "Occlusion" when all the surfaces of the carton are occluded, and "All" when at least one surface of the carton is not occluded. To segment different surfaces of the carton, we have designed a new labeling method as shown in Figure \ref{fig:label} and the \textbf{labeling rules} includes: (1) only the two-line points can be selected as the start points; (2) the points should be labeled in a clockwise direction; (3) the common lines should be repeatedly labeled twice.
  
\subsection{Surfaces segmentation algorithm}
\label{subS:3.3}
  \textbf{Data procssing:} Based on our labeling method, it is easy to judge whether the point is a faceted point according to definition \ref{df1}. But during labeling process, the coordinates of the same faceted points have errors when labeling according to the third labeling rule. Thus, we judge the point is the same faceted points in labeled points set $P^S$(as shown Figure \ref{fig:label}) according to Eq.\ref{eq1}(3 to 11 line in Algorithm\ref{alg:r2p}). Then the average of the coordinate about the same faceted points is used to replace the values of the same faceted points in $P^S$ to get a new point set as $P^{new}$(17 to 18 line in Algorithm\ref{alg:r2p}). Finally, we get a point set $P^{t}$ without repeated points from $P^{new}$(12 to 16 line in Algorithm\ref{alg:r2p}) to calculate the cost matrix.

\newtheorem{definition}{\textbf{Definition}}
\begin{definition}
The three-line points are faceted points as shown in Fig.\ref{fig:label}. According to the labeling rules in section \ref{subS:3.2}, the points which are repeated more than 1 times in the labeled points are faceted points.
\label{df1}
\end{definition}

\begin{equation}
\label{eq1}
{D(p_i,p_j)}=\sqrt{(p_{i_x}-p_{j_x})^2+(p_{i_y}-p_{j_y})^2}\leq\psi
\end{equation}
  In Eq.\ref{eq1}, $p_i,p_j\in P^S$, and $(p_{i_x},p_{i_y})$ is the coordinate of $p_i$. $\psi$ is a hyperparameter to determine whether two points are the same faceted point,$\psi=25$ here.
  
  \begin{algorithm}[]
    \caption{Surfaces segmentation algorithm }  
    \LinesNumbered 
    \label{alg:r2p}  
    \KwIn{$P^S=\{p_i\}_{i=0}^n$: the points list of the instance from labeling points;}  
    \KwOut{$P_{result}=\{p_i\}_{i=0}^2$ :the points of surfaces in the carton;}  
    \textbf{Initialize:} {$\psi=25$;\quad$index=[[],[]...]$: the Array which save the indexes of points with same coordinate in $P^s$;\quad $P^t=[]$:the Array which do not have repeated points.\\
    }
    \textbf{\#Data procssing:}\\
    \For{$i=0$ to n }
      {  
       $index{[i]}\gets \{i\}$\\
       \For{$j=0$ to n}
        {
         \text{Calculate $D(P^S[i],P^S[j])$ by Eq.(\ref{eq1})} \\
         \If{$D(P^S[i],P^S[j])\le \psi$}
            {  \eIf{$i$ is $j$}{continue}{$index{[i]}\gets index{[i]}+\{j\}$}
            } 
        }
      }
    $index_{rm} \gets RemoveDuplicateData(index)$\\
    $K \gets Length(index_{rm})$ \\
    \For {$i=0$ to $K$}
     {
       $N \gets Length(index_{remove}[i])$\\
       $P^t[i] \gets \frac{\sum_{k=0}^NP^S[index_{rm}[i][k]]}{N}$\\
       \For{$j=0$ to $N$}
       {
         $P^{new}[index_{rm}[i][j]]\gets P^t[j]$
       }
     } 

   \textbf{\#Computational cost matrix $V$:} \\
   \For{ $i=0$ to $K$ }
     {
       \For{ $j=0$ to $K$ }
       {
         \eIf{$i$ is $j$}
         {
           continue
         }{
           calculate $V_{[i][j]}$ by Eq.(\ref{eq2}) with $P^t[i],P^t[j]$ and $P^{new}$
         }
        }
      }
   \textbf{\#The TSP solution:} \\
     $D_s=[]$ \\
     \For{$k=0$ to $K$}
      {
           $N \gets Length(index_{rm}[k])$\\
           \eIf{$N >1$}
           {
             continue
           }{
               $p_{init} \gets P^t[k]$\\
               $\overline{P}_{/k}\gets P^t-\{P^t[k]\}$\\
               \text{get shortest closed polygons as} $D_{s}{[i]}$ by Eq.(\ref{eq3}) with $V, \overline{P}_{/k}$ and $p_{init}$
            }
      }
      ${P_{result}\gets RemoveDuplicateSurface(D_s)}$
\end{algorithm}
  
  \textbf{Calculating cost matrix:} For Traveling Salesman Problem(TSP), it is important to get a cost matrix about each city. In our method, each labeled point is regarded as a city. We suppose that the $G(P^t,V)$ is a directed graph, $V=[V_{i,j}]_{k\times k}$ is the cost matrix between each point in $P^t$ which is calculated by Eq.\ref{eq2}(20 to 25 line in Algorithm\ref{alg:r2p}). In Eq.\ref{eq2}, the symbol \textbf{$\rightarrow$} means the index of point $p_i$ in $P^{new}$ plus 1 is equal to the index of $p_j$ in $P^{new}$. It means that $p_i$ to $p_j$ is connected and assigned the value 1, otherwise it is not connected and the value is infinite represented by $Inf$.

\begin{equation}
\label{eq2}
{V_{i,j}}={\left\{\begin{aligned}
                    & 1, p_i \rightarrow p_j\\
                    & Inf\\
                   \end{aligned}
            \right.}
        \qquad\textbf{s.t}{\left\{\begin{aligned}
        & p_i,p_j\in P^t\\
        & P^t \subseteq P^{new}\\
             \end{aligned}
              \right.}
\end{equation}

  \textbf{TSP algorithm:} 
  The surfaces segmentation is equivalent to find all closed polygons without overlap in $G(P^t, V)$, and we can use the TSP\cite{gavish1978travelling} method to find the closed polygons. Firstly, we set two-line point as an initial point $p_{init}$(28 to 33 line in Algorithm\ref{alg:r2p}). Secondly we remove the initial point in $P^{t}$ and replace it with $\overline{P}$(34 line in Algorithm\ref{alg:r2p}). Then the path back to $p_{init}$ in $\overline{P}$ calculated with the least cost by Eq.\ref{eq3} is the segmented surface. Finally, each point of two-line points as initial point is used to get the surfaces and remove the duplicate surfaces(28 to 36 line in Algorithm\ref{alg:r2p}).
  
\begin{equation}
\label{eq3}
{F(p_{init},G(\overline{P},V))}=\min{F(p_k,G(\overline{P}_{/k},V))+V_{init,k}}
\quad \text{s.t} \left\{\begin{aligned}
       &  p_{init}\in P^t\\
       &  p_k\in \overline{P}\\
       &  \overline{P}_{/k}=\overline{P}-\{p_k\}\\
       & \sum_{i=0}^{|P^t|}V_{i,j}\leq 3\\
       &  \sum_{j=0}^{|P^t|}V_{i,j}\leq 3\\
       \end{aligned}
       \right.
\end{equation}
In Eq.\ref{eq3}, $\overline{P}_{/k}=\overline{P}-\{p_k\}$ is the unsearched point set. $\sum_{i=0}^{|P^t|}V_{i,j}\leq 3$ and $\sum_{j=0}^{|P^t|}V_{i,j}\leq 3$ are the conditions that any point in $P^t$ is connected with other points at most 3 times. And when {$V_{k,init}=1$}, the $F(p_{k},G(\overline{P},V))$ ends up and the path is a closed polygon. Finally the path with least cost is a surface of the carton.
  
\subsection{Contour reconstruction algorithm}
\label{subS:3.4}
\textbf{Theoretical analysis:} As we know, for an incomplete contour of surface, it can be obtained by occlusion and truncation with many parallelograms, and for parallelograms with too large scale, it is easy to cause the synthesized image to be not realistic. So we add an additional condition with smallest area when we construct a complete contour. Assuming that the incomplete contour of the surface is $X$ and the parallelogram contour is $Y$, the goal is to find a optimal parallelogram $Y^{\ast}$ with our conditions in contour sets ${\Omega}^{R^2}$, as shown in Eq.\ref{eq5}.

\begin{equation}
\label{eq5}
{Y^{\ast}}=\mathop{\arg\max}_{Y\in {\Omega}^{R^2}}P(Y|X)
\quad \text{s.t} \left\{\begin{aligned}
        & S_{Y^{\ast}}=\min S_Y\\
        & K_{A}=K_{C} \And K_{B}=K_{D} \\
       \end{aligned}
       \right.
\end{equation}
In Eq.\ref{eq5}, $S_{Y^*}=\min S_Y$ represents the complete contour with the smallest area. $K$ is the slope of the line, $A,B,C$,and $D$ are the edge of parallelogram. $K_{A}=K_{C}$ and $K_{B}=K_{D}$ are the conditions of the parallelogram rule to construct a parallelogram.

Because of the perspective transformation principle, some surfaces with complete contours in $X$ may not satisfy the parallelogram contour. In order to maintain the perspective transformation relationship of the complete contour as much as possible, we compare the area of the original complete contour(it is assumed that the contour represented by 4 points is a complete contour) with the area of the parallelogram constructed by Eq.\ref{eq5} to get the final contour. The final contour $Y_{final}$ is calculated in Eq.\ref{eq6}.
\begin{equation}
\label{eq6}
{Y_{final}}=\left\{\begin{aligned}
                    & y ,\qquad\textbf{s.t} {\left\{\begin{aligned}
                              {\gamma<\frac{area(y)}{area(Y^*)} }\\
                              len(y) = 4\\
                              \end{aligned}
                              \right.}\\
                    & Y^*,\qquad\qquad\qquad others\\
                   \end{aligned}
            \right.
\end{equation}
In Eq.\ref{eq6}, $y$ represents the original contour. The $len(y)=4$ is the condition that the original contour is represented by 4 points. $area(*)$ is the area of each contour. $\gamma$ is the hyperparameter to decide the final contour and $\gamma=2/3$ is used in our experiments .

\textbf{Parallelogram reconstruction:} As we all known, the parallelograms satisfy the convex polygon condition(as shown in definition\ref{df2}). So during constructing a parallelogram, we select a line $A$($ap_{i_x}+bp_{i_y}+c=0$) from the edge of incomplete contour as the constructed line(as shown in definition\ref{df3}). And let $P_{surface}=\{p_0,p_1,…,p_n \}$ denote the incomplete contour. When $A$ and it's adjacent edge $B$ in $P_{surface}$ both satisfy the convex polygon condition, we construct a parallelogram with $A$ and $B$. Firstly, we get the points $P_A$ with farthest distance to $A$, and $P_B$ with farthest distance to $B$ in $P_{surface}$. Then we calculate the slope of $A$ and $B$ as $K_A$, $K_B$ and construct the other edges($C$ and $D$) of parallelogram with ($K_A$,$P_A$) and ($K_B$,$P_B$)(22 to 25 line in algorithm\ref{alg:r3p}). Finally, we calculate the point of intersection of each edge in $A,B,C,D$ to get the parallelogram(26 to 29 line in algorithm\ref{alg:r3p}).
\begin{definition}
 Convex polygon condition: All points in the parallelogram must be on the side of any edge.
\label{df2}
\end{definition}
\begin{definition}
 Constructed line: when constructing a parallelogram, the selected line which satisfy the convex polygon condition is the edge of the parallelogram is a constructed line. It is judged by Eq.\ref{eq7} and Eq.\ref{eq8}. And when $|\beta|$ equals to $|m-n|$, the selected line is a constructed line(8 to 18 line in algorithm\ref{alg:r3p}).
\label{df3}
\end{definition}
\begin{equation}
\label{eq7}
\beta = \sum_{i=0}^n\beta_i\\
\quad \text{s.t} \left\{\begin{aligned}
     & p_i \in P_{surface}\\
      & \beta_i={\left\{\begin{aligned}
      & 1 , \qquad if: ap_{i_x}+bp_{i_y}+c>0\\
      & 0 , \qquad if: ap_{i_x}+bp_{i_y}+c=0\\
      & -1, \qquad\qquad\qquad others\\
      \end{aligned}
       \right.}\\
       \end{aligned}
       \right.
\end{equation}

\begin{equation}
\label{eq8}
m=\sum_{i=0}^{n}g_i\
\quad \text{s.t}\quad  
      g_i={\left\{\begin{aligned}
      & 1,\qquad if: \beta_i=0\\
      & 0,\qquad others
       \end{aligned}
       \right.
      }\\
\end{equation}
In Eq.\ref{eq7} and Eq.\ref{eq8}, $m$ is the number of points in the constructed line. $(p_{i_x},p_{i_y})$ is the coordinate of point $p_i$ in $P_{surface}$.

\begin{algorithm}
\caption{single visible surface reconstruction}  
\LinesNumbered 
\label{alg:r3p} 
\KwIn{$P_{surface}=\{p_i\}_{i=0}^n$: incomplete contour points; \quad$Label$: label information }  
\KwOut{the complete contour points of surface: $P_{c}=\{p_i\}_{i=0}^3$}  
\textbf{Initialize:}{\quad$FLAG=False$: True means the constructed line and it's adjacent edge both satisfy the convex condition;\quad$(A,B,C)$ is the parameters of line $Ax+By+C=0$.\\}
\eIf{ $Label$ is $all$}
{
 $P_c \gets P_{surface}$\\
}
{
 \For{$i=0$ to n}
 {
  calculate the parameters  $(A_1,B_1,C_1)$ of the $line_1$ with $P_{surface}[i],P_{surface}[i+1]$\\
  calculate the parameters  $(A_2,B_2,C_2)$ of the $line_2$ with $P_{surface}[i+1],P_{surface}[i+2]$\\
   \For{$k=0$ to $n$}
   {
    $\beta_1[k],g_1[k]$ $\gets$ calculated by Eq.(\ref{eq7}) and Eq.(\ref{eq8}) with $line_1$ and $P_{surface}[k]$\\
    $\beta_2[k],g_2[k]$ $\gets$ calculated by Eq.(\ref{eq7}) and Eq.(\ref{eq8}) with $line_2$ and $P_{surface}[k]$\\
   }
  $\beta_{line1} \gets \sum_{i=0}^n\beta_1[i]$\\
  $m_{line1} \gets \sum_{i=0}^ng_1[i]$\\
  $\beta_{line2} \gets \sum_{i=0}^n\beta_2[i]$\\
  $m_{line2} \gets \sum_{i=0}^ng_2[i]$\\
  \eIf{$|\beta_{line1}|$ is $|n-m_{line1}|$ and $|\beta_{line2}|$ is $|n-m_{line2}|$}{
  $FLAG \gets True$ }{$FLAG \gets False$}
  $K_A \gets \frac{P_{surface}[i]_y-P_{surface}[i+1]_y}{P_{surface}[i]_x-P_{surface}[i+1]_x}$ \qquad\#slope of line1\\
  $K_B \gets \frac{P_{surface}[i+1]_y-P_{surface}[i+2]_y}{P_{surface}[i+1]_x-P_{surface}[i+2]_x}$\qquad\#slope of line2\\
  \eIf{FLAG is True}
   {
   $P_{A}$ is the point with farthest distance to line1 in $P_{surface}$\\
   $P_{B}$ is the point with farthest distance to line2 in $P_{surface}$\\
  calculate the parameters  $(A3,B3,C3)$ of the $line_3$ with $K_A$ and $P_{A}$\\
  calculate the parameters  $(A4,B4,C4)$ of the $line_4$ with $K_B$ and $P_{B}$\\
  $Point[0]\gets P_{surface}[i+1]$\\
  $Point[1]\gets CalculatePointofIntersection(line_3,line_2)$\\
  $Point[2]\gets CalculatePointofIntersection(line_3,line_4)$\\
  $Point[3]\gets CalculatePointofIntersection(line_1,line_4)$\\
  $P_{list[i]}\gets Point$\\
  $S_{[i]} \gets area(Point)$ \\
   }
{
   $P_{list[i]} \gets Null$\\
   $S_{[i]} \gets Infinity$ \\
}
}
\eIf{$n$ is not $4$}
{$P_c$ $\gets$ Get the complete contour by Eq.(\ref{eq5}) with $P_{list}$ and $S$}
{$P_c$ $\gets$ Get the complete contour by Eq.(\ref{eq5}) and Eq.(\ref{eq6}) with $P_{list}$ and $S$}
}
\end{algorithm}
\textbf{Parallelogram  reconstruction strategy:}
Due to the perspective transformation of imaging, the cartons appear in three kinds of appearance in the image including single visible surface, two visible surfaces and three visible surfaces. The initial condition of the contour reconstruction methods for these three kinds of appearance are as follows: 

(1) {single visible surface reconstruction:}  When there is no occlusion as \textbf{All} according to the labeling rules, the contour keeps unchanged. When the carton is labeled as \textbf{Occlusion}, we select each edge in $P_{surface}$ as the constructed line to construct a parallelogram and we choose the parallelogram with smallest area as the contour of the $P_{surface}$(as shown in algorithm\ref{alg:r3p}).

(2) {two visible surfaces:} Because there is a common line, when we construct a parallelogram, we only use the common line as constructed line to construct parallelograms for each surface. And we also choose the parallelogram with smallest area as the contour of the $P_{surface}$. Finally, for the parallelogram of each surface, we should adjust it's scale, so that the coordinate of the common line in each surface is equal in image.

(3) {three visible surfaces:} Because there are two common lines in each surface, when we construct a parallelogram, the common lines as constructed line and its adjacent edge are used to construct a parallelogram for each surface. Finally, for the parallelogram of each surface, we also should adjust its scale, so that the coordinate of the common line in each surface is equal in image.

\subsection{Image synthesis}
\label{subS:3.5}
\begin{figure}[]
\centering
\subfloat[]{\label{Fig6a}
\includegraphics[width=0.45\linewidth]{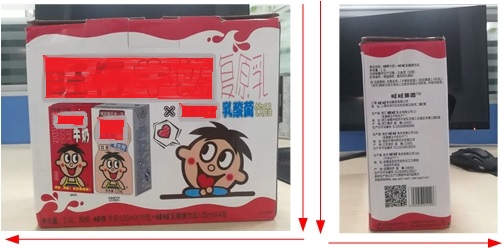}}
\subfloat[]{\label{Fig6b}
\includegraphics[width=0.3\linewidth]{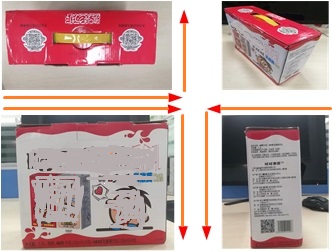}}
\caption{The example of foreground texture labeling rules. (a) is the labeling rules of the two visible surfaces, which started to the common line; (b) is the labeling rules of the three visible surfaces. As we can see in the first row and last column, the combination relationship of the surface is fixed with each other in the carton instance.}
\label{fig6}
\end{figure}

\textbf{Foreground texture datasets:} We have collected different kinds of cartons which have only one surface as the foreground texture datasets(as shown in Fig.\ref{Fig:2d}). Because the carton is composed of multiple surfaces and each surface has a specific combination relationship(such as the direction of the surface) with each other. The method shown in Figure \ref{fig6} is used to label the foreground texture to ensure the true relationship with each other. And we build a subset in the foreground texture datasets according to the number of visible surfaces.

\textbf{Image synthesis method:} During performing the image synthesis algorithm, Perspective transformation principle is used to adaptively deform the foreground instance to ensure the linearity of the edges. The coordinates of the foreground instance are $P_{pre}=\{p_i\}_{i=0}^3$, and the coordinates of the reconstructed contour are $P_{back}=\{p_j\}_{j=0}^3$, the perspective principle is as follows:
\begin{equation}
\label{eq9}
\left[
  \begin{array}{c}
    P_{pre}\\
    1\\
  \end{array}
\right]=M\times\left[\begin{array}{c}   
    P_{back}\\
    1\\
  \end{array}
  \right]=\left[                 
  \begin{array}{ccc}   
    a_{00} &a_{01} & a_{02}\\
    a_{10} &a_{11} & a_{12}\\
    a_{20} &a_{21} & a_{22}\\
  \end{array}
\right]\left[                 
  \begin{array}{c}   
    P_{back}\\
    1\\
  \end{array}
\right] 
\end{equation}
 $M$ can be solved when $P_{pre}$ and $P_{back}$ are fixed. Then, $M$ is used to generate images by Eq.\ref{eq10}. In Eq.\ref{eq10}, $I$ is the original image from source datasets, $I_{pre}$ is the foreground instance texture from target datasets, $I_{back}$ is the mask of the reconstructed contour, $I_x$ is the mask of the original instance in $I$ and $I_{synthetic}$ is the synthetic image. The $\odot$ represents the image fusion operation \cite{dwibedi2017cut} as shown in the fifth module in Figure \ref{fig:3}. And $\oplus$ represents the pixel-level image merging operation.
\begin{equation}
\label{eq10}
I_{synthetic}=(M\times I_{pre}\odot I_x)\oplus((\textbf{1}-I_x)\odot I)\\
\end{equation}

 Because the carton is composed of multiple surfaces in the image, the texture from foreground texture datasets needs to be selected according to the relationship of the surface in $I$. For single visible surface, $I_{pre}$ is randomly selected. And for the two visible surfaces and three visible surfaces, we firstly select the subset of the foreground dataset corresponding to the number of surfaces, then we randomly select the foreground instance textures to generate image.

\textbf{Subtle pixel artifacts:} In the process of image synthesis, because of the brightness difference between $I_{pre}$ and $I$, there must be some artificial noise in $I_{synthetic}$, such as the subtle pixel artifacts at the edge of the synthesized instance\cite{dwibedi2017cut}. In order to reduce the influence of subtle pixel artifacts, we use Gaussian fusion method for foreground texture fusion. And the random noise texture is used to replace the instance texture as the background with a probability of 0.2 which isn't labeled(as shown in Figure \ref{fig:3}) to make the detection model focus only on the object appearance.

\section{Experiments}
\label{S:4}
This chapter mainly explores the application of surfaces segmentation algorithms and contour construction algorithms in the field of data expansion. We verify the effectiveness of our method on the object detectors such as RetinaNet\cite{lin2017focal} and Faster R-CNN\cite{ren2016faster}. All experiments are based on PyTorch and conducted on 2 GTX1080Ti.

\subsection{Experimental Settings}
\label{S:4.1}
\textbf{Datasets:} The information of the SCD is shown in Table\ref{SCDinformation}. There are 520 images from CPLC as the carton stacking skeleton and used the method in chapter\ref{subS:3.2} to label them. Then, we collect 269 single-sided instances, 51 double-sided instances, and 23 three-sided instances from FM as the foreground texture datasets. In addition, 149 single-sided instances, 27 double-sided instances, and 30 three-sided instances are collected from ECLC as the other foreground texture datasets. All experiments take CPLC as a base dataset which is split into 3589 images for training and 500 images for testing.

\begin{table}[htbp]
    \vspace{10pt}
    \centering
    \begin{tabular}{c c c c c c c }
        \hline
        \multirow{2}*{\text{Scene}} &\multirow{2}*{\text{Train No.}} &\multirow{2}*{\text{Test No.}} &\multirow{2}*{\text{Stacking skeleton No.}}& \multicolumn{3}{c}{foreground texture datasets}\\
         ~&  ~ & ~ & ~ & Single sides & Two sides & three sides\\
        \hline
        \text{CPLC} & 3589 & 500 & 520 & 0 & 0 & 0\\
        \text{ECLC} & 1722 & 500 & 0 & 149 & 27 & 30\\
        \text{FM} & 0 & 492 & 0 & 269 & 51 & 23 \\
        \hline       
    \end{tabular}
    \caption{Data distribution of the SCD datasets in different scenarios and distribution of the texture datasets}
    \label{SCDinformation}
\end{table}

 \textbf{Implementation details:} All experiments are implemented on basis of MMDetection \cite{chen2019mmdetection}. We utilize ResNet-18 in RetinaNet\cite{lin2017focal} as backbones which are pre-trained on ImageNet. A mini-batch of 4 images per GPU is used during training RetinaNet\cite{lin2017focal} and Faster R-CNN\cite{ren2016faster}, thus making a total mini-batch of 8 images on 2 GPUs. The synchronized Stochastic Gradient Descent (SGD) is used for model optimization. The weight decay of 0.0001 and the momentum of 0.9 are adopted. A linear scaling rule\cite{goyal2017linear} is carried out to set the learning rate during training (0.005 in RetinaNet and 0.01 in Faster R-CNN). And a linear warm-up strategy is adopted in the first 500 iterations. Besides that the learning rate changes linearly with mini-batch, the flip ratio is 0.5 and the image scale is [600, 1000] in all experiments. And other settings are consistent with the default settings of MMDetection\cite{chen2019mmdetection}.
 
\begin{figure}[htbp]
\centering
\includegraphics[height=0.45\linewidth]{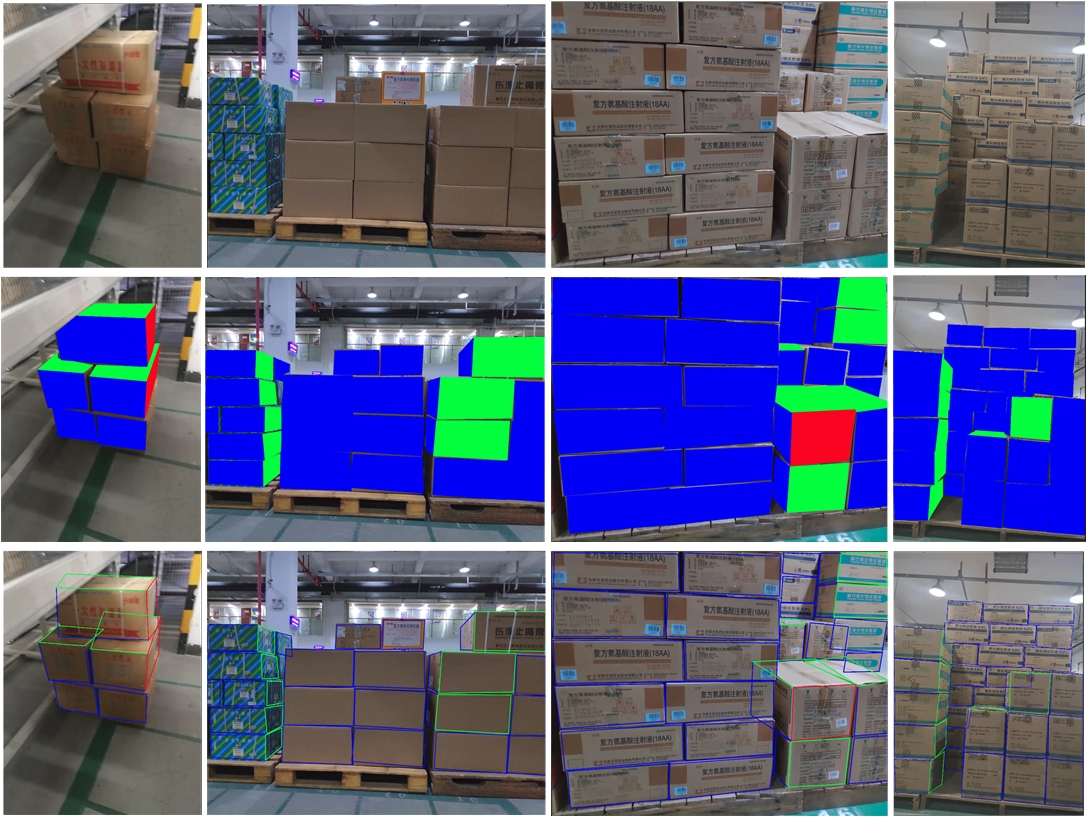}
\caption{The results of surface segmentation algorithm and contour reconstruction algorithm. On the second row, the blue, green, and red is the different texture of the carton. On the third row, the quadrilateral with different colors is the complete contour.}
\label{fig:7}
\end{figure}
 \subsection{Qualitative Analysis}
\label{S:4.2}
\begin{figure}[htbp]
\centering
\includegraphics[height=0.45\linewidth]{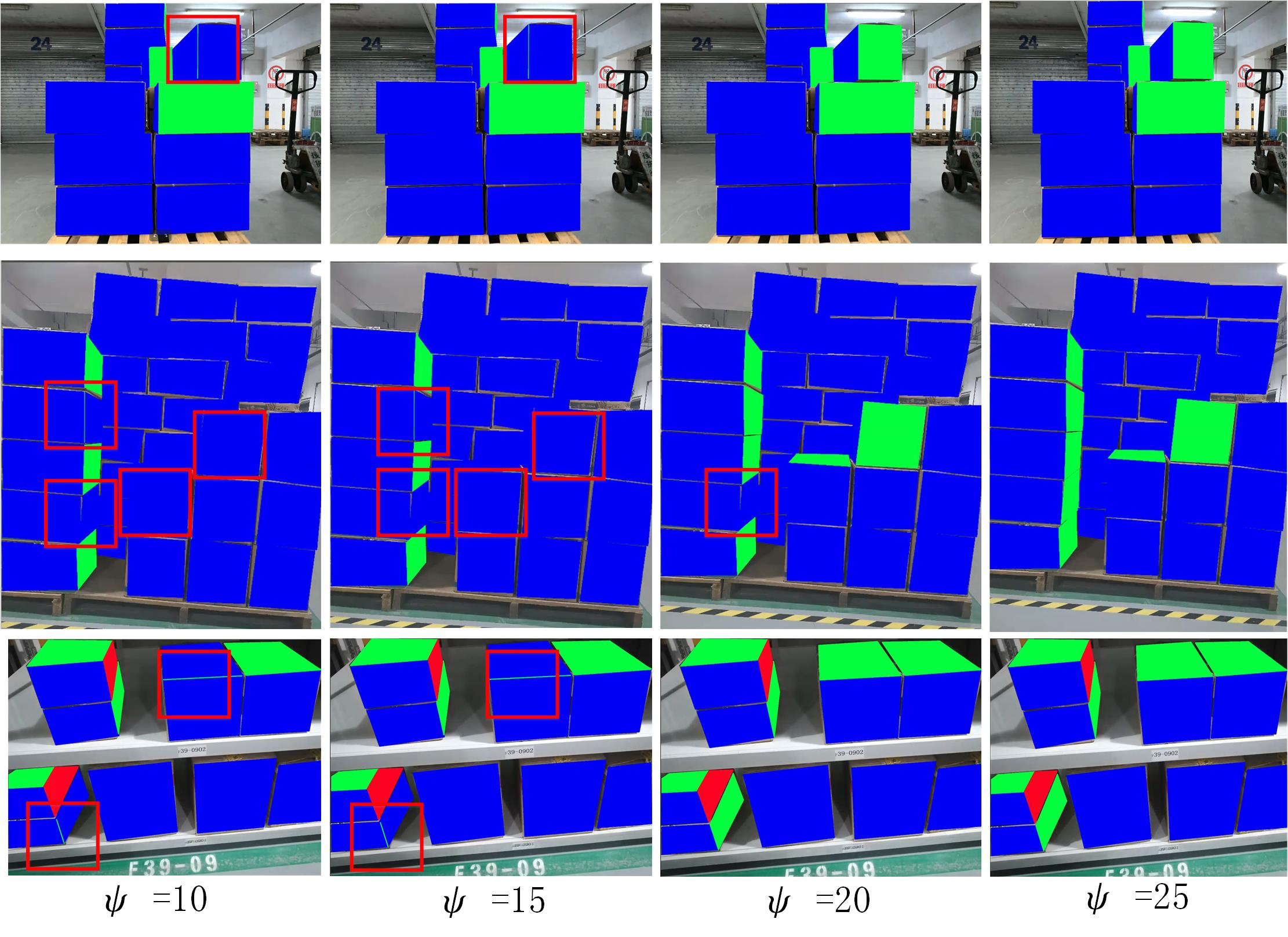}
\caption{The results of surfaces segmentation algorithm with different hyperparameters $\psi$. The blue, green, and red are the different texture of the carton. The red square is the result of an error.}
\label{fig:8}
\end{figure}

\textbf{The results of surfaces segmentation algorithm:} We selects 520 images as the carton stacking skeleton datasets from CPLC and then surfaces segmentation algorithm are used to decouple the texture of the skeleton(as shown in the second row of Fig.\ref{fig:7}). It can completely decouple the texture of the carton.  And sampling statistics are used to search the optimal hyperparameter $\psi$. For each group of the $\psi$, we randomly select 10 pictures to decouple the texture and repeated five times. Then we compute the error rate of the surfaces segmentation. Table \ref{paramterfacted} shows that the best result is founded when $\psi=25$, and Fig.\ref{fig:8} shows the corresponding visualization result.
\begin{table}[htbp]
    \vspace{10pt}
    \centering
    \begin{tabular}{c c c c c c c }
        \hline
         ~ &1 &2 &3 &4 &5 &average\\
        \hline
        $\psi=10$ &3.52\% &2.8\% &0 &9.41\% &8.33\% &4.81\%\\
        $\psi=15$ &1.41\% &2.8\% &0 &1.34\% &5.00\% &2.11\%\\
        $\psi=20$ &0      &0     &0 &0      &0.08\% &0.016\%\\
        $\psi=25$ &0      &0     &0 &0      &0      &0\\
        \hline       
    \end{tabular}
    \caption{The error rate of the surfaces segmentation with different hyperparameters $\psi$.}
    \label{paramterfacted}
\end{table}

\begin{figure}[]
\centering
\subfloat[]{\label{Fig9a}
\includegraphics[width=0.45\linewidth]{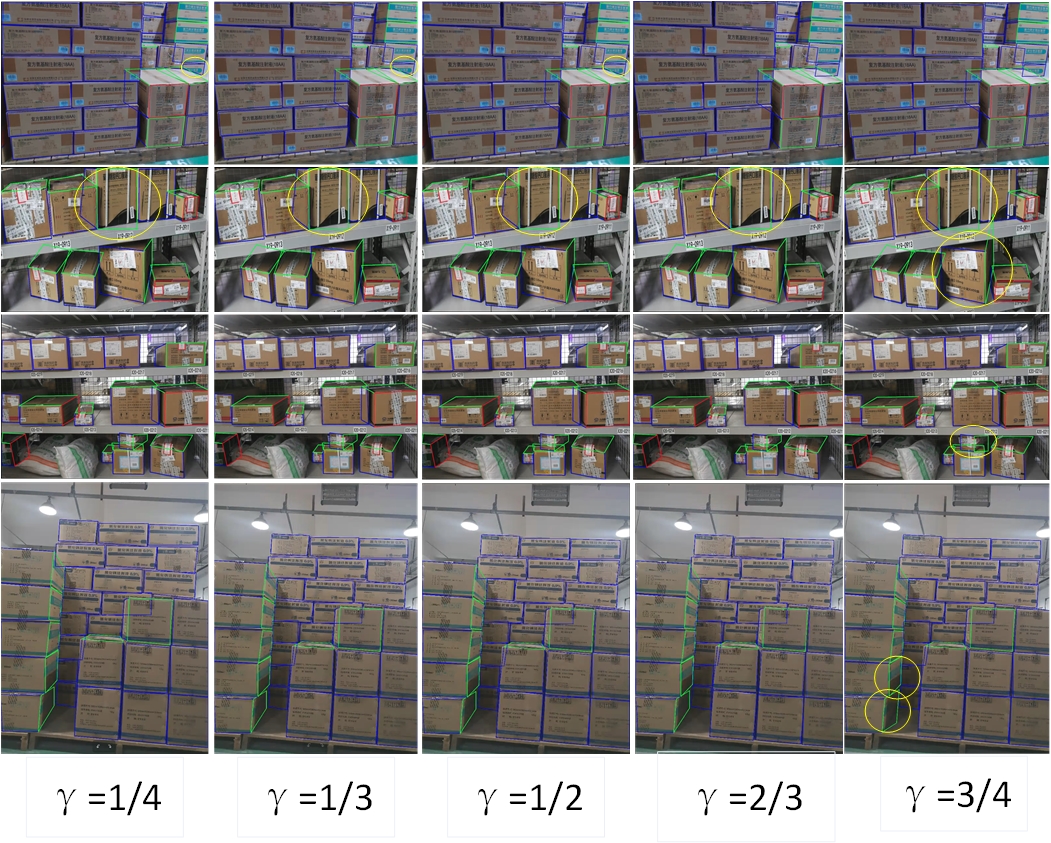}}
\subfloat[]{\label{Fig9b}
\includegraphics[width=0.45\linewidth]{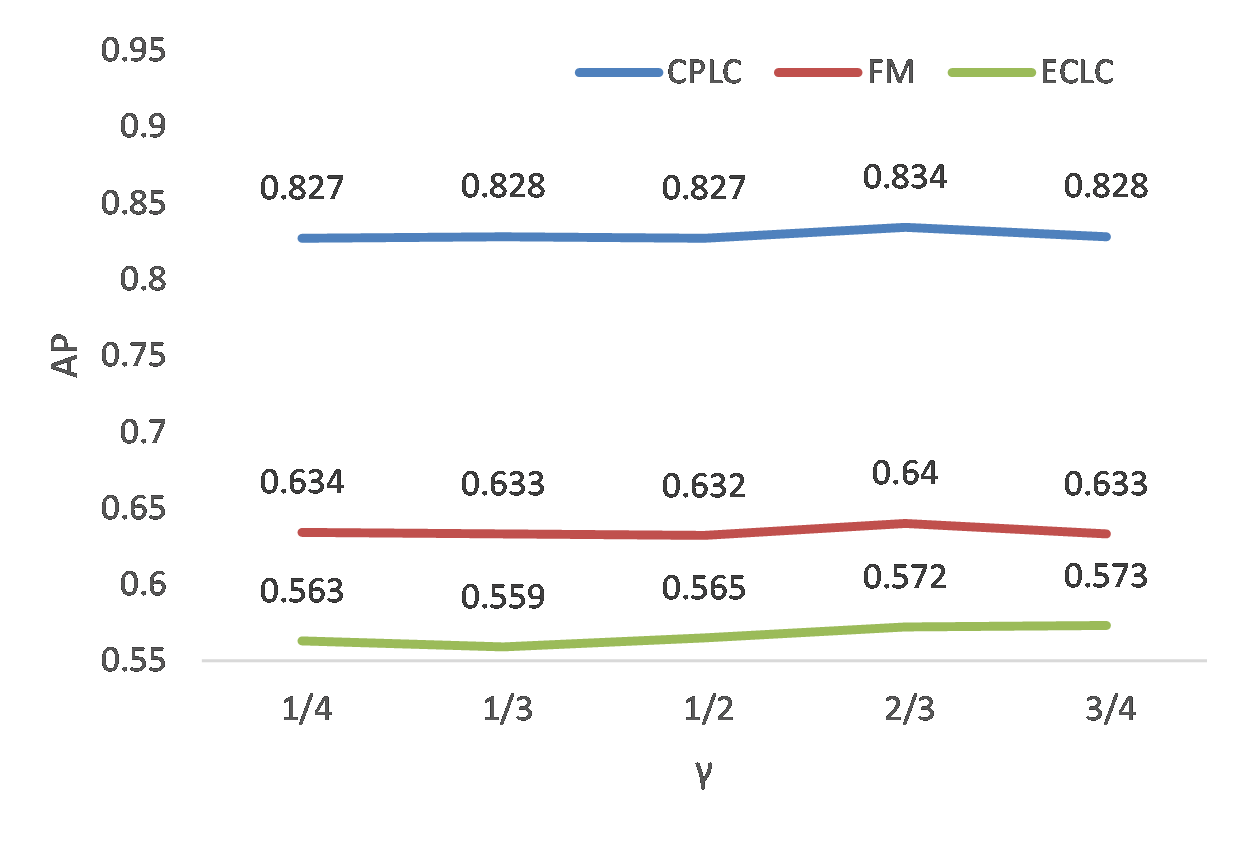}}
\caption{(a) the results of the contour reconstruction with different $\gamma$. The yellow ellipse is the unreasonable result. (b) The results of the RetinaNet\cite{lin2017focal} model trained by the generated images with different $\gamma$.}
\label{fig:9}
\end{figure}

\textbf{The results of contour reconstruction algorithm:} The result of the contour reconstruction algorithm is effective and reasonable as shown in the third row of Fig.\ref{fig:7}. To get the optimal value of the $\gamma$, we randomly select 10 pictures for each group of $\gamma$ to construct the complete contour and repeated five times. Finally, we compute the percentage of unreasonable samples (the examples are occluded but not reconstructed or reconstructed but unreasonable by comparison with the original complete contour). The best parameter of $\gamma$ is $\frac{2}{3}$ as shown in Table\ref{constrct}, while Fig.\ref{Fig9a} shows the corresponding visualization result. We also use the FM as texture to generate 6000 images for each $\gamma$  and mix them with the CPLC, then train the RetinaNet. As shown in Fig.\ref{Fig9b}, it is the same with the result of the sampling statistics.
\begin{table}[htbp]
    \centering
    \begin{tabular}{c c c c c c c }
        \hline
         ~ &1 &2 &3 &4 &5 &average\\
        \hline
        $\gamma=\frac{1}{4}$ &9.86\% &10.19\% &8.72\% &8.24\% &10.83\% &9.57\% \\
        $\gamma=\frac{1}{3}$ &9.86\% &9.26\%  &8.72\% &8.24\% &10.00\% &9.22\% \\
        $\gamma=\frac{1}{2}$ &9.86\% &9.26\%  &8.72\% &8.24\% &10.00\% &9.16\% \\
        $\gamma=\frac{2}{3}$ &7.04\% &8.33\%  &6.71\% &8.24\% &9.17\%  &7.90\% \\
        $\gamma=\frac{3}{4}$ &7.75\% &9.26\%  &7.38\% &8.24\% &9.17\%  &8.36\% \\
        \hline       
    \end{tabular}
    \caption{The ratio of unreasonable contours generated by different values of the $\gamma$ to the total number of sample instances}
    \label{constrct}
\end{table}

\begin{figure}[htbp]
\centering
\subfloat[]{\label{fig:10}
\includegraphics[width=0.45\linewidth]{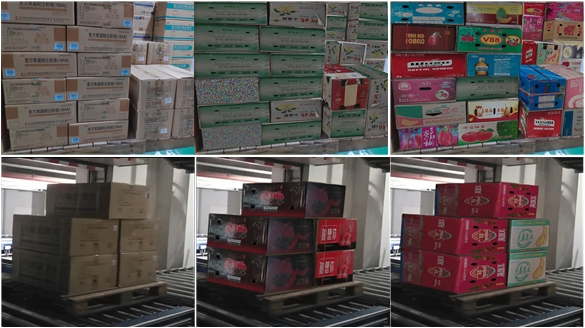}}
\subfloat[]{\label{fig:11}
\includegraphics[width=0.45\linewidth]{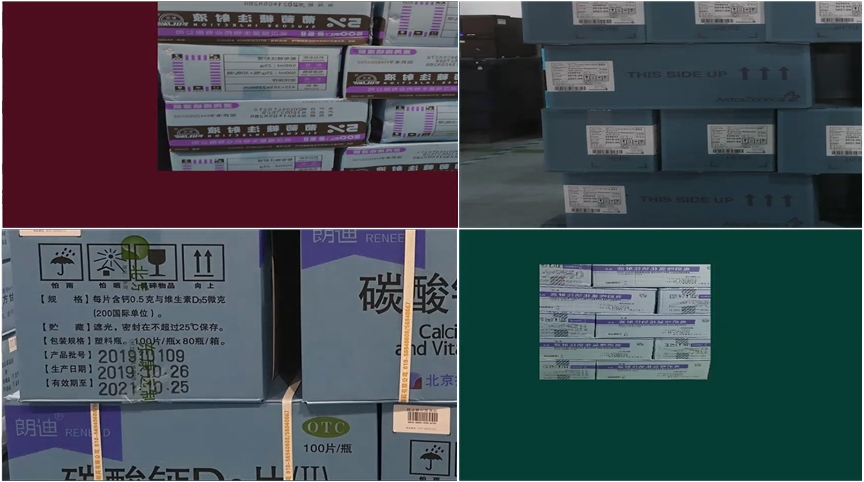}}
\caption{(a) The first column is the skeleton pictures and the rest are the synthesized images by our method. (b) Data augmentation for CPLC with random scaling, random flipping, random cutting, random expanding, etc.}
\end{figure}
\textbf{Image synthesis:} The Gaussian fusion method is used to generate images. In the processing of image synthesis, the foreground texture datasets are randomly selected according to the number of the surfaces on the skeleton picture. And then we replace the instance texture on the skeleton picture with the texture from foreground texture datasets, as shown in \ref{fig:10}. And a random noise is used to solve the influence of the subtle pixel artifacts, and the noise texture isn't labeled. And in the Fig.\ref{fig:10}, our synthesized images looks more real compared with Fig.\ref{fig:1}, because the context relationship of carton instance and background is reasonable.

\subsection{Main Results}
\label{S:4.3}
\textbf{Baselines on carton datasets:} For the baselines, RetinaNet\cite{lin2017focal} and Faster R-CNN\cite{ren2016faster} equipped with ResNet18 are used to fine-tuning on the training set of CPLC and test respectively on 500 images from ECLC and 492 images from FM. The overall results are reported in Table \ref{SGCMP_E} which shows that huge domain shift exists from CPCL to FM and ECLC, up to $23\% \sim 26.8\%$ in AP.

\begin{table}[htbp]
\vspace{12pt}
\centering
\begin{tabular}{c c c c c c c}
\hline
{datasets} & {train No.} & {model} & {test} & $AP$ & $AP_{50}$ & $AP_{80}$\\
\hline
\multirow{3}*{\text{Baseline: CPLC}} & \multirow{3}*{3589} & \multirow{9}*{\text{RetinaNet{\cite{lin2017focal}}}} & \text{CPLC} & \text{0.814} & \text{0.970} & \text{0.877}\\
~ & ~ & ~& \text{FM} & \text{0.575} & \text{0.810} & \text{0.569}\\
~ & ~ & ~& \text{ECLC} & \text{0.546} & \text{0.798} & \text{0.520} \\

\multirow{3}*{\text{CPLC+AUG}} & \multirow{3}*{9589} & ~ & \text{CPLC} & \text{0.829} & \text{0.969} & \text{0.8895}\\
~ & ~ & ~& \text{FM} & \text{0.588} & \text{0.812} & \text{0.589}\\
~ & ~ & ~& \text{ECLC} & \text{0.552} & \text{0.791} & \text{0.533} \\
\multirow{3}*{\text{CPLC+G\_FM}} & \multirow{3}*{9589} & ~ & \textbf{CPLC} & \textbf{0.834} & \textbf{0.973} & \textbf{0.896} \\
~ & ~ & ~& \textbf{FM} & \textbf{0.640} & \textbf{0.852} & \textbf{0.650} \\
~ & ~ & ~& \textbf{ECLC} & \textbf{0.572} & \textbf{0.810} & \textbf{0.568} \\
\hline
\multirow{3}*{\text{Baseline: CPLC}} & \multirow{3}*{3589} & \multirow{9}*{\text{Faster R-CNN\cite{ren2016faster}}}& \text{CPLC}  & \text{0.837} & \text{0.976} & \text{0.902}\\
~ & ~ & ~& \text{FM} & \text{0.607} & \text{0.811} & \text{0.624}\\
~ & ~ & ~& \text{ECLC} & \text{0.577} & \text{0.813} & \text{0.580} \\
\multirow{3}*{\text{CPLC+AUG}} & \multirow{3}*{9589} & ~ & \text{CPLC}  & \text{0.846} & \text{0.976} & \text{0.910}\\
~ & ~ & ~& \text{FM} & \text{0.616} & \text{0.815} & \text{0.625}\\
~ & ~ & ~& \text{ECLC} & \text{0.573} & \text{0.792} & \text{0.573} \\
 
\multirow{3}*{\text{CPLC+G\_FM}} & \multirow{3}*{9589} & ~ & \textbf{CPLC} & \textbf{0.852} & \textbf{0.977} & \textbf{0.914} \\
~ & ~ & ~& \textbf{FM} & \textbf{0.675} & \textbf{0.855} & \textbf{0.700} \\
~ & ~ & ~& \textbf{ECLC} & \textbf{0.593} & \textbf{0.806} & \textbf{0.588} \\
\hline
\end{tabular}
\caption{The performance of the method in this paper to generate FM foreground texture images on the detection model.}
\label{SGCMP_E}
\end{table}
\textbf{Foreground texture datasets from FM:} 343 special foreground texture from FM are plugged into 520 skeletons extracted from CPLC. Then we replace the texture of each instance on the skeleton images with the foreground texture and a random noise texture with a probability of 0.2. Finally, we used the Gaussian fusion method to generate 6000 images and mix them with CPLC datasets as \text{CPLC+G}\_{FM}. For comparison, we use the data augmentation method to generate 6000 images (as shown in Figure \ref{fig:11}) as AUG which are subsequently mixed with CPLC as  \text{CPLC+AUG}. The comparative experiments are shown in Table \ref{SGCMP_E}.

As shown in Table \ref{SGCMP_E}, when RetinaNet and Faster R-CNN are trained on \text{CPLC+G\_FM} and tested on CPLC, the AP can be improved by 2\% and 1.5\% by comparison with training on CPLC. In addition, compared with only training on \text{CPLC+AUG}, the performance of our method can improve AP by 0.5\% and 0.6\%. This demonstrates that our method is better than the traditional data augmentation strategy. And when RetinaNet and Faster-RCNN are tested on FM, the AP increases by 6.5\% and 6.8\% by comparison with training on CPLC. When testing on ECLC, the AP also can be improved by 2.6\% and 1.6\% by comparison with training on CPLC. This demonstrates that our method can greatly ease the domain shift. Finally, RetinaNet and Faster R-CNN are tested on FM and ECLC, the performance of our method can improve AP by 5.2\% and 5.9\% on FM test set and 2\% and 2\% on ECLC test set, compared with training on \text{CPLC+AUG}. This demonstrates that our method is not only better than the traditional data augmentation strategy for data augmentation, but also can effectively ameliorate the domain shift.

\begin{table}[htbp]
\vspace{12pt}
\centering
\begin{tabular}{c c c c c c c}
\hline
{datasets} & {train No.} & {model} & {test} & $AP$ & $AP_{50}$ & $AP_{80}$\\
\hline
\multirow{3}*{\text{Baseline: CPLC}} & \multirow{3}*{3589} & \multirow{9}*{\text{RetinaNet{\cite{lin2017focal}}}} & \text{CPLC} & \text{0.814} & \text{0.970} & \text{0.877}\\
~ & ~ & ~& \text{FM} & \text{0.575} & \text{0.810} & \text{0.569}\\
~ & ~ & ~& \text{ECLC} & \text{0.546} & \text{0.798} & \text{0.520} \\

\multirow{3}*{\text{CPLC+AUG}} & \multirow{3}*{9589} & ~ & \text{CPLC} & \text{0.829} & \text{0.969} & \text{0.8895}\\
~ & ~ & ~& \text{FM} & \text{0.588} & \text{0.812} & \text{0.589}\\
~ & ~ & ~& \text{ECLC} & \text{0.552} & \text{0.791} & \text{0.533} \\

\multirow{3}*{\text{CPLC+G\_ECLC}} & \multirow{3}*{9589} & ~ & \textbf{CPLC} & \textbf{0.831} & \textbf{0.972} & \textbf{0.894} \\
~ & ~ & ~& \textbf{FM} & \textbf{0.595} & \textbf{0.820} & \textbf{0.598} \\
~ & ~ & ~& \textbf{ECLC} & \textbf{0.589} & \textbf{0.812} & \textbf{0.575} \\
\hline
\multirow{3}*{\text{Baseline: CPLC}} & \multirow{3}*{3589} & \multirow{9}*{\text{Faster R-CNN\cite{ren2016faster}}}& \text{CPLC}  & \text{0.837} & \text{0.976} & \text{0.902}\\
~ & ~ & ~& \text{FM} & \text{0.607} & \text{0.811} & \text{0.624}\\
~ & ~ & ~& \text{ECLC} & \text{0.577} & \text{0.813} & \text{0.580} \\

\multirow{3}*{\text{CPLC+AUG}} & \multirow{3}*{9589} & ~ & \text{CPLC}  & \text{0.846} & \text{0.976} & \text{0.910}\\
~ & ~ & ~& \text{FM} & \text{0.616} & \text{0.815} & \text{0.625}\\
~ & ~ & ~& \text{ECLC} & \text{0.573} & \text{0.792} & \text{0.573} \\

\multirow{3}*{\text{CPLC+G\_ECLC}} & \multirow{3}*{9589} & ~ & \textbf{CPLC} & \textbf{0.846} & \textbf{0.975} & \textbf{0.909} \\
~ & ~ & ~& \textbf{FM} & \textbf{0.624} & \textbf{0.830} & \textbf{0.635} \\
~ & ~ & ~& \textbf{ECLC} & \textbf{0.611} & \textbf{0.818} & \textbf{0.627} \\
\hline
\end{tabular}
\caption{The performance of the method in this paper to generate ECLC foreground texture data on the detection model which compared with data augmentation methods}
\label{JDCMP_E}
\end{table}

\textbf{Foreground texture datasets from ECLC:} The foreground instances in ECLC which contain 206 instances is used as foreground texture datasets, the skeleton in CPLC as the template datasets. And the parameters of all algorithms are kept the same as before. The Gaussian fusion method is used to generate 6000 images, which are mixed with CPLC datasets as \text{CPLC+G\_ECLC}, the results are shown in Table \ref{JDCMP_E}.

It can be seen in Table \ref{JDCMP_E} that the AP of the \text{CPLC+G\_ECLC} is improved by 1.7\% on CPLC by comparison with the baseline on the RetinaNet. And the AP of RetinaNet training on \text{CPLC+G\_ECLC} is improved by 2\% on the FM, 4.3\% on ECLC by comparison with training Baseline. At the same time, compared with the \text{CPLC+AUG}, the \text{CPLC+G\_ECLC} has a slightly improvement of only 0.2\% on the CPLC, 0.7\% on FM, but it is improved by 3.7\% on ECLC. In the Faster R-CNN, our method improve by 0.9\% on CPLC, 1.7\% on FM, and 3.4\% on ECLC by comparison with baseline. Our method outperforms the \text{CPLC+AUG} with 0.8\% on FM and 3.8\% on test sets of CPLC. This further demonstrates that our method is not only better than the traditional data augmentation strategy for data augmentation, but also can effectively solve the domain shift. And the number of foreground texture datasets from ECLC is smaller than that from FM, so it causes the performance of ECLC to be lower than that of FM. 

\subsection{Ablation Studies}
\begin{figure}[htbp]
\centering
\subfloat[]{\label{Fig15a}
\includegraphics[width=0.4\linewidth]{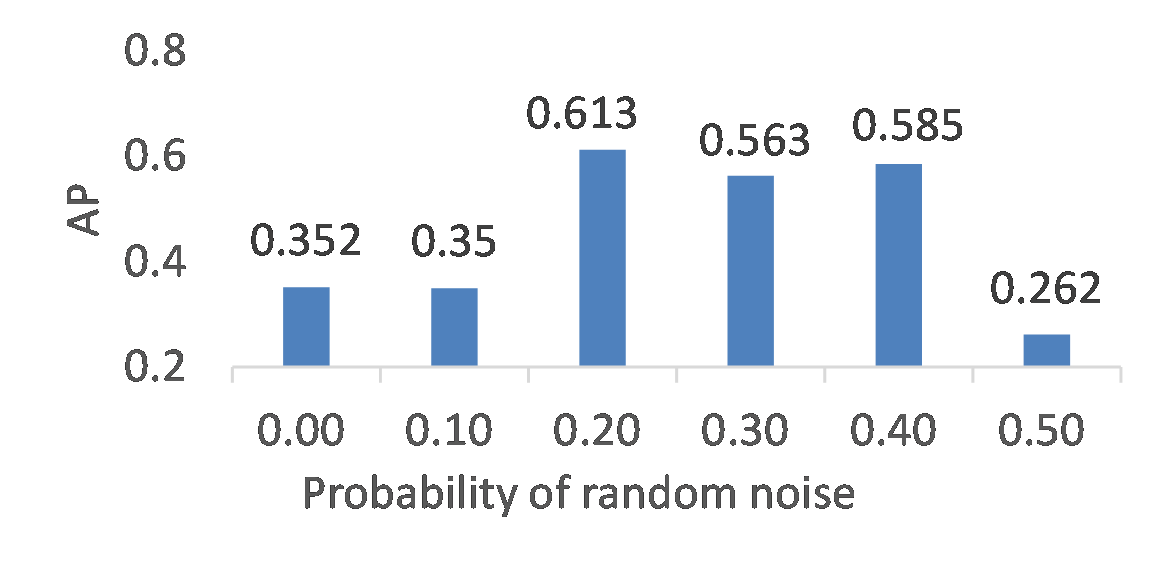}}
\subfloat[]{\label{Fig15b}
\includegraphics[width=0.4\linewidth]{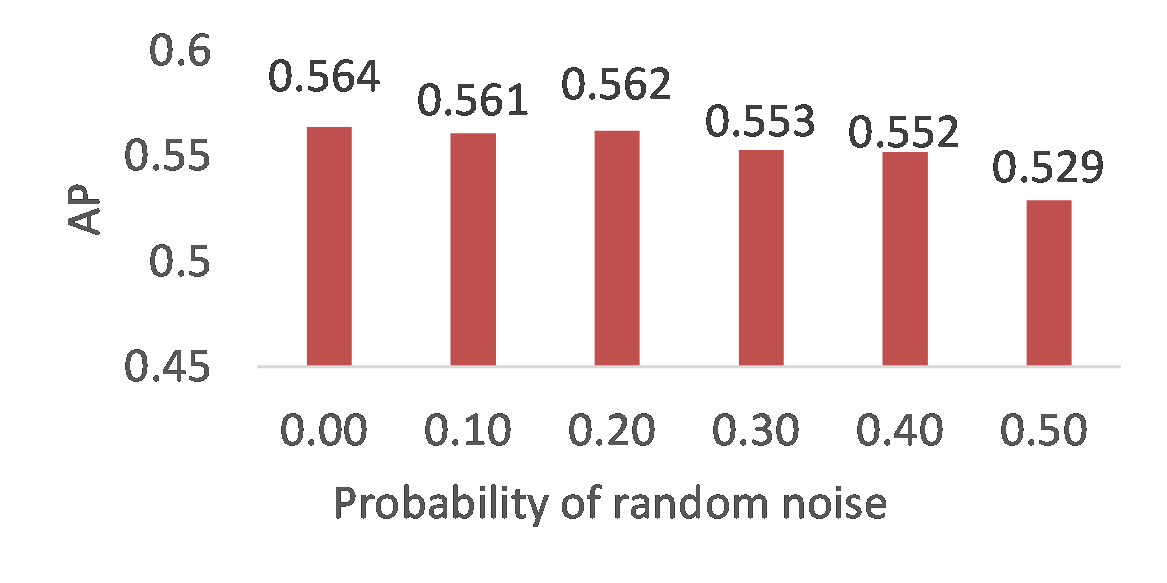}}\\
\subfloat[]{\label{Fig15c}
\includegraphics[width=0.4\linewidth]{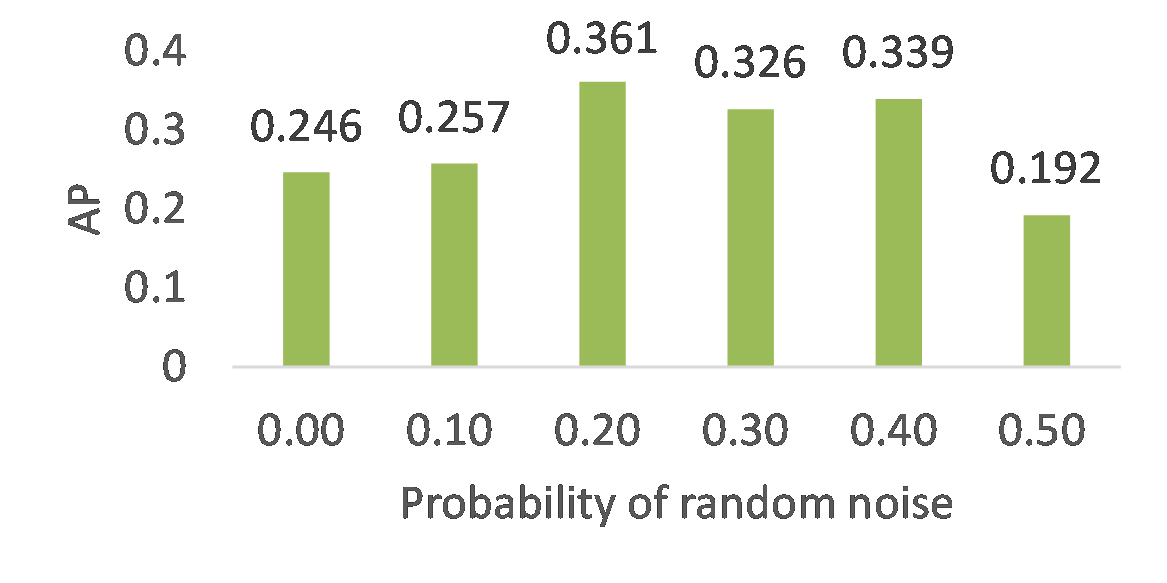}}
\caption{The influence of the probability on RetinaNet that the noise texture is the foreground instance texture. (a) is the result of testing on CPLC,(b) is the result of testing on FM,(c) is the result of testing on ECLC. The best probability is 0.2.}
\label{fig:15}
\end{figure}

\textbf{The influence of the probability on RetinaNet that the noise texture is the foreground texture:} To explore the impact of random noise for the detectors, the foreground texture dataset from FM is used as a benchmark to generate 3589 images with different probabilities of the random noise to train the RetinaNet\cite{lin2017focal} and test on the CPLC, FM, and ECLC test sets. According to the results in Figure \ref{fig:15}, the performance of the model is the best on CPLC and ECLC test sets when the probability of the random noise is 0.2, which means that the random noise can effectively effectively suppress the influence of artificial noise on the model and make the detectors focus only on the object appearance.

\begin{figure}[htbp]
\centering
\includegraphics[width=0.5\linewidth]{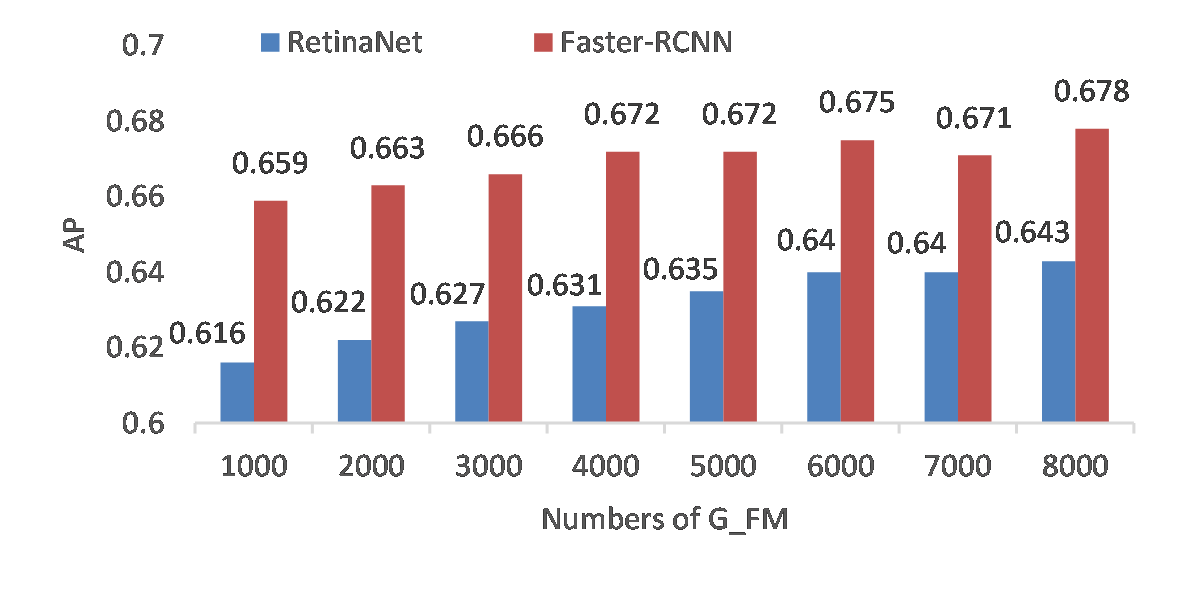}
\caption{The influence of the number of the synthesized images on detectors. FM is the foreground texture datasets to generate different numbers of training sets as G\_FM. And the G\_FM is mixed with CPLC to train the detection model and test them on the FM test sets.}
\label{fig:14}
\end{figure}
\textbf{The influence of the number of the synthesized images on detectors:} To explore the influence of the number of the synthesized images on the detectors, FM is used as the foreground instance texture datasets and the random noise probability is set as 0.2 to generate different numbers of images. Then they are mixed with the CPLC training datasets to train the RetinaNet \cite{lin2017focal} and Faster R-CNN \cite{ren2016faster} and tested on FM. As shown in Figure \ref{fig:14}, the performance of the models increases as the number of the generated images. Compared with the baseline in Table \ref{SGCMP_E}, the performance of our methods is improved by 6.8\% on RetinaNet and 7.1\% on Faster-RCNN when the number of synthesized images is 8000. However, the performance of the RetinaNet and Faster R-CNN increase slowly when the number of the synthesized images exceeds 6000. Because the influence of subtle pixel artifacts, our method is saturated when the number of synthesized images is more than 6000.
\begin{figure}[htbp]
\centering
\subfloat[]{\label{Fig12a}
\includegraphics[width=0.35\linewidth]{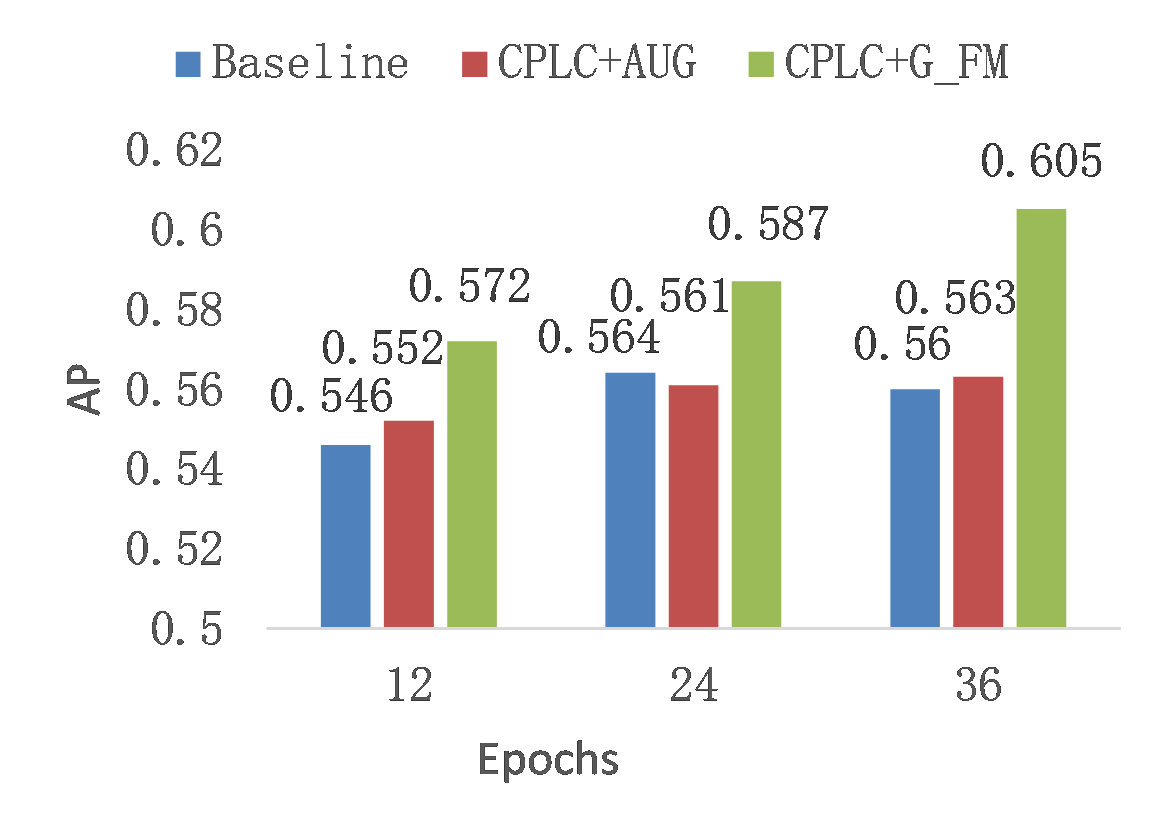}}
\subfloat[]{\label{Fig12b}
\includegraphics[width=0.35\linewidth]{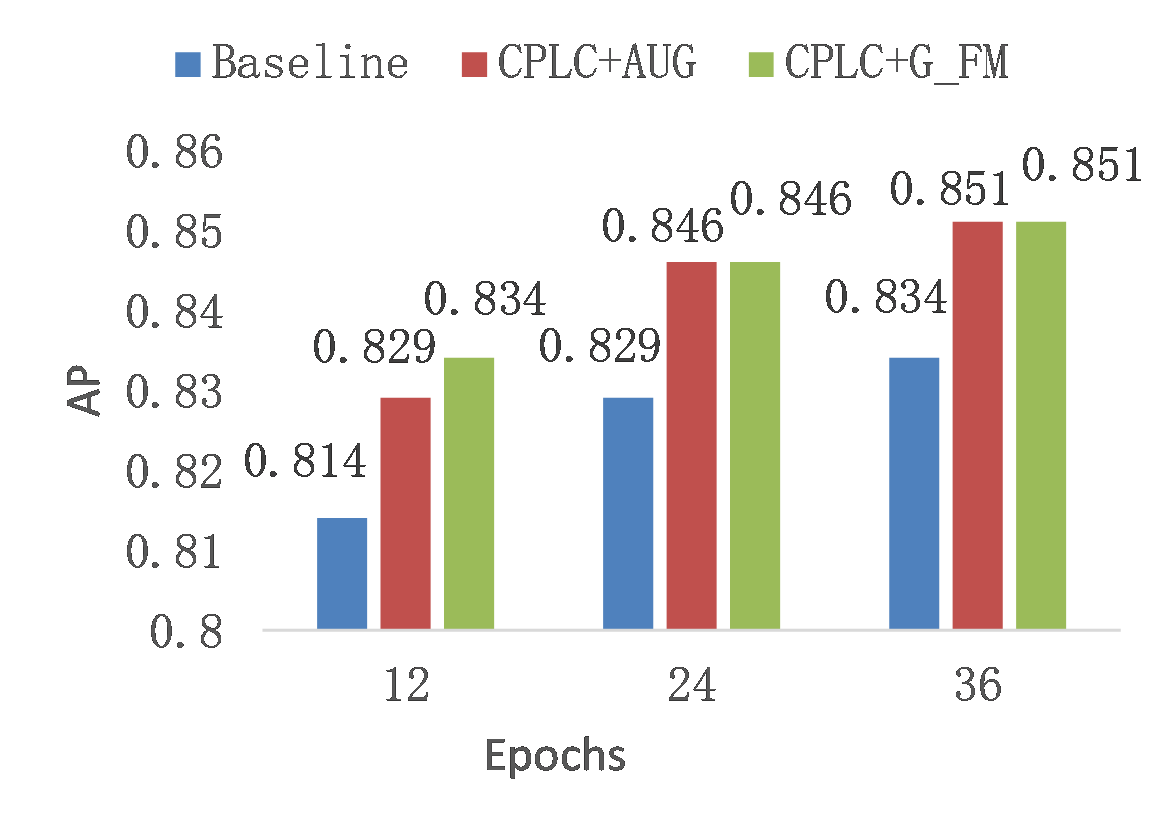}}

\subfloat[]{\label{Fig12c}
\includegraphics[width=0.35\linewidth]{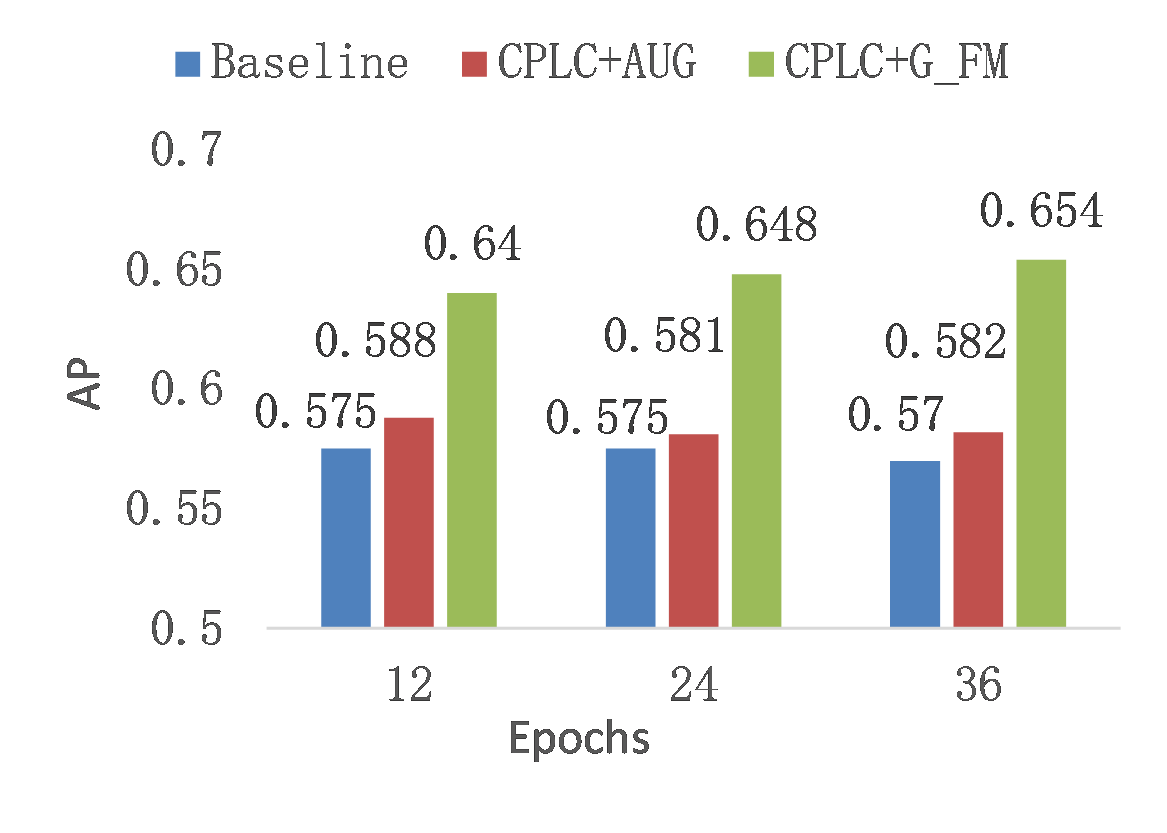}}
\caption{The influence of epochs in RetinaNet \cite{lin2017focal}, (a) is the result of testing on the ECLC test sets, (b) is the result of testing on the CPLC test sets, (c) is the result of testing on FM test sets.}
\label{fig:12}
\end{figure}

\textbf{The influence of training epochs in RetinaNet:} We do experiments to explore the influence of the number of training epochs on the RetinaNet(as shown in Figure \ref{fig:12}). The \text{CPLC+G\_FM} datasets in Chapter \ref{S:4.3} are used as the training datasets. The performance of our method on FM test set by comparison with the Baseline is improved to 8.4\% when the number of training epochs is 36. It is proved that our method can learn more texture information by increasing the training time. And when testing on CPLC, our method is the best on 12th training epochs, on the other training epochs, our method can keep the same performance compared with CPLC+AUG. Because the number of epochs increases, the model learns more basic features and the contribution of the context relationship is saturated for CPLC. 

\section{Conclusions and future work}
\label{S:5}
 In this work, we propose a data synthesis approach to quickly synthesize annotated training images and address the issue of cross-domain carton detection. Our method can completely decouple the texture of the carton, and the foreground texture replacement of the carton can make the synthesized images looks more real. And our method not only increase the performance of the source domain but also the target domain. At the same time, the random noise as foreground texture can reduce the harm of the subtle pixel artifacts to the model.

 In the future, we will build a larger common carton stacking skeleton dataset. And we will explore the application of this framework in one-shot learning, synthesized images brightness adaptive, and how to further reduce the harm of the subtle pixel artifacts to the model. 

\section{Acknowledgements}
\label{S:6}
This research did not receive any specific grant from funding agencies in the public, commercial, or not-for-profit sectors. 

\bibliographystyle{elsarticle-num-names}
\bibliography{sample.bib}
\end{document}